\documentclass[a4paper,11pt,oneside]{book}

\usepackage[margin=1.2in]{geometry} 

\usepackage{titlesec}
\usepackage{amsfonts}

\usepackage[cmex10]{amsmath}
\usepackage{amssymb}
\usepackage{amsthm}
\newtheorem{definition}{Definition}
\newtheorem{theorem}{Theorem}
\numberwithin{definition}{chapter}
\numberwithin{theorem}{chapter}

\usepackage{algorithm}
\usepackage{algpseudocode}

\usepackage{placeins}
\usepackage[utf8]{inputenc}
\usepackage{listings}
\usepackage{xcolor}
\definecolor{codegreen}{rgb}{0,0.6,0.1}
\definecolor{codegray}{rgb}{0.5,0.5,0.5}
\definecolor{codeblue}{rgb}{0.10,0.00,1.00}
\definecolor{codepurple}{rgb}{0.58,0,0.82}
\definecolor{backcolour}{rgb}{1.0,1.0,1.0}

\lstdefinestyle{mystyle}{
    backgroundcolor=\color{backcolour},   
    commentstyle=\color{codegreen},
    keywordstyle=\color{codeblue},
    numberstyle=\tiny\color{codegray},
    stringstyle=\color{codepurple},
    basicstyle=\ttfamily\footnotesize,
    breakatwhitespace=false,         
    breaklines=true,                 
    captionpos=b,                        
    keepspaces=true,                 
    numbers=left,                    
    numbersep=5pt,                  
    showspaces=false,                
    showstringspaces=false,
    showtabs=false,                  
    tabsize=2,
    frame=none
}
\usepackage{graphicx}
\usepackage{caption}
\usepackage{lipsum}

\usepackage{multirow}
\usepackage{rotating}
\usepackage{makecell}
\usepackage{booktabs}
\usepackage{comment}
\usepackage[bottom]{footmisc}
\usepackage{subfig}
\usepackage{tikz,lipsum,lmodern}
\usepackage[most]{tcolorbox}

\newtcolorbox{tbluebox}[1]{
    colback=blue!5!white,
    colframe=cyan!75!black,
    title=#1
}

\newtcolorbox{tredbox}{
    colback=red!5!white,
    colframe=red!75!black
}

\usepackage{enumitem}
\newlist{abbrv}{itemize}{1}
\setlist[abbrv,1]{label=,labelwidth=1in,align=parleft,itemsep=0.1\baselineskip,leftmargin=!}

\usepackage[hidelinks]{hyperref}

\usepackage{biblatex} 
\usepackage[toc]{appendix}
\begin{document}

    \captionsetup[figure]{margin=1.5cm,font=small,name={Figure},labelsep=colon}
    \captionsetup[table]{margin=1.5cm,font=small,name={Table},labelsep=colon}

    \frontmatter
    
    \begin{titlepage}      
        \begin{center}
			
            \linespread{1}~\\[3cm]\Huge {
                \textbf{Neural Network Learner for Minesweeper}
            
            }
            \linespread{1}~\\[1cm]

            {\LARGE Loughborough University\\[0.5cm]
            Department of Computer Science}\\[2cm]
            
            {\Large 
                Muhammad Hamza Sajjad
            }\\[1cm]

            {\large 
                \emph{Supervisor:} Dr Daniel Reidenbach}\\[1cm] 
            
            \large The report submitted as fulfilment of the requirements of\\the Loughborough University for the degree of\\ Bachelor of Science in \textit{Computer Science}\\[0.3cm] 
            \vfill

            \textbf{}{May 3, 2022} 
        \end{center}
    \end{titlepage}

    
\chapter*{\center \Large  Abstract}


Minesweeper is an interesting single player game based on logic, memory and guessing. Solving Minesweeper has been shown to be an NP-hard task. Deterministic solvers are the best known approach for solving Minesweeper. This project proposes a neural network based learner for solving Minesweeper. To choose the best learner, different architectures and configurations of neural networks were  trained on hundreds of thousands of games. Surprisingly, the proposed neural network based learner has shown to be a very good approximation function for solving Minesweeper. The neural network learner competes well with the CSP solvers, especially in Beginner and Intermediate modes of the game. It was also observed that despite having high success rates, the best neural learner was considerably slower than the best deterministic solver. This report also discusses the overheads and limitations faced while creating highly successful neural networks for Minesweeper.

    \chapter*{\center \Large  Acknowledgements}
I would like to thank Dr.\ Daniel Reidenbach for his interest and support throughout the project. His guidance and feedback has been very helpful for a successful completion of this project. I would also like to thank David Becerra and Tai-Yen Wu for providing the CSP solvers from their projects. David Becerra's CSP solver with its source code has been critical for speed comparison between the learners and the CSP solver.


    
    \tableofcontents
    
    \mainmatter
    
    \chapter{Introduction}
\label{ch:into} 

Minesweeper is a board game containing hidden mines with clues about the positions of the mines. The combination of logic, different difficulty levels and sometimes luck make it a very interesting game to play. As we shall see, solving Minesweeper is an NP-hard problem and creating a solver for this game with high success rates is a challenging task. There are different deterministic approaches used to solve Minesweeper that have been discussed in the literature.

This project will focus on implementing an AI learner for Minesweeper game with the aim to achieve fairly high success rates. Multiple learners from different machine learning techniques will be used and compared in terms of success rates. The central question of this project is whether or not highly successful machine learning models can be created for solving Minesweeper.

\section{Motivation}
\label{sec:into_motiv}
Many interesting problems like Minesweeper are difficult and fall into the class of NP-hard problems. One approach for solving these problems are SAT solvers which are applied after reducing the given problem to an instance of the Satisfiability problem. Example of some other approaches are identifying heuristics for the problem which work reasonably well in many cases and restricting the size of the problem by making it simpler and solvable for most of cases of interest.

With the amount of research and progress being witnessed in the field of AI, the question is how well an AI learner will perform when trained for solving Minesweeper. There are many other board games and video games such as Go, Chess, Super Mario and attacking queens problem which have been learned by machine learning techniques with good approximations\footnote{"Good approximation" is dependent on the type of the game. For example, AI learners for Go and Chess beating the world champions is considered to be a good approximation.}. When a machine learning enthusiast faces the problem of defining a solver for such a problem, machine learning techniques would be their first choice because these techniques have been successfully tested on various board games and video games.

For many problems, AI approaches have been successful because computers are able to brute force over big search spaces allowing AI models to learn patterns for solving these challenging problems. Can the power of brute force combined with an AI solve Minesweeper efficiently?

This project will also reflect on the other point of view that criticizes the extensive usage of AI for problems that can be modelled by deterministic solvers. Many computer scientists would argue that we should not overuse AI approaches for problems that can be modelled by algorithms, Minesweeper is certainly one of those problems.

\section{Aims and objectives}
\label{sec:intro_aims_obj}

\textbf{Aims:} The aim of this project is to create an AI learner that can reach high success rates, ideally as high as deterministic solvers which are known to be the best approach for solving Minesweeper. If a learner with high success rate is not found, this gives us an opportunity to reflect on why the deterministic solvers are a better approach. 

\noindent \textbf{Objectives:} 
\begin{itemize}
    \item Understand Minesweeper: the first starting step of the project will be to get an understanding of the Minesweeper game. This includes a study of its rules, different difficulty levels and identifying approaches that are used by a player for solving Minesweeper.
    \item Literature review: This includes understanding how difficult is Minesweeper from the complexity theory point of view. Before designing a solver it is necessary to find existing solvers discussed in the literature and their success rates. The literature review will be important for setting the success rate from the literature as a goal that the learners should achieve.
    \item Implementation of Minesweeper: an implementation of the Minesweeper game will be needed for training the learner. This can either be a reuse of an open source version of the game available online or a custom implementation of the game from scratch.
    \item Implementing and training the learner: this will be the most important and time-consuming part of the project. The successful achievement of this objective will allow us to analyze different learners. This objective will involve in researching different AI techniques that are suitable for Minesweeper. A game play strategy will be needed for each type of learner which will dictate how the game will be played. The implementation and training of the learner may overlap because usually a model needs to be tweaked several times before it can reach acceptable success rates. 
    \item Evaluation of the experiments: the training of the network will be followed by a thorough analysis of the results produced by the learners. This will involve in discussing the success rates of the learners on different difficulty levels of the game. The combined results of different learning models will allow us to derive conclusions about which technique is more suitable.
\end{itemize}

    \chapter{Background and Literature Review}
\label{ch:lit_rev} 

\section{Introduction to Minesweeper}
\label{ch:intro_to_minesweeper}
Minesweeper became popular after being pre-installed on Windows OS from 1990 and onward \cite{minesweeper_wiki}. It is a single player puzzle game that typically consists of a board of n x m cells. The board contains mines hidden in the cells and each cell not containing a mine carries a number indicating how many adjacent cells have a mine. In the initial configuration all the cells on the board are covered and the player wins when all non-mine cells are revealed. If the player uncovers a cell containing mine, the mine explodes and the game is lost.

\begin{figure}[ht] 
    \centering
    \subfloat[Initial configuration]{%
        \includegraphics[width=0.45\textwidth]{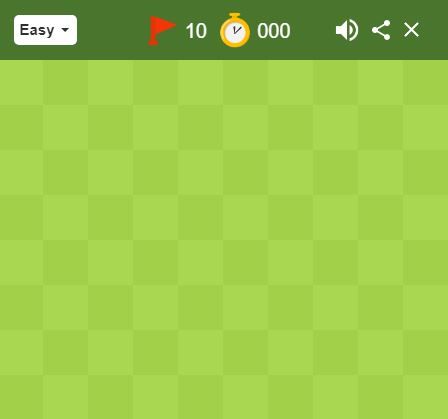}%
        \label{fig:minesweeper_init_config}%
        }%
    \hfill%
    \subfloat[Playing configuration]{%
        \includegraphics[width=0.45\textwidth]{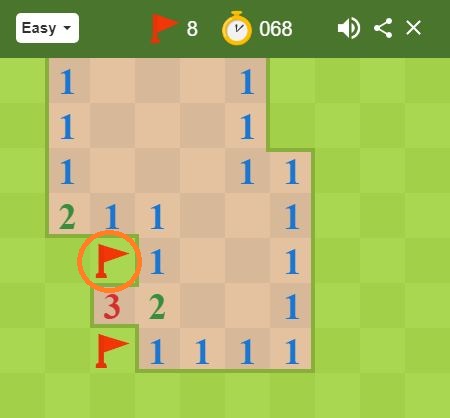}%
        \label{fig:minesweeper_second_config}%
        }%
    \caption{Minesweeper Game by Google}
    \label{fig:minesweeper_play_configs}
\end{figure}

As shown in Figure~\ref{fig:minesweeper_init_config}, in the initial configuration of the game all the cells are covered and the user is shown the number of mines in this board which depends on the difficulty level. In the Figure~\ref{fig:minesweeper_second_config} there is a game configuration after the player has made some moves. Each non-mine tile contains a clue about its neighbours indicating exactly how many mines do they contain. The empty cell can be seen as containing a zero because none of its neighbours have a mine. The neighbours of a cell are all the cells adjacent to it vertically, horizontally and diagonally. Hence a non-mine cell contains a number between 1 and 8 or is empty if it has no mines in its neighbours.

The Figure~\ref{fig:minesweeper_second_config} also shows cells marked with flags. When the player identifies that a tile carries a mine, they can mark it with a flag. This is not a mandatory move, but it helps the player avoiding memorizing all the cells that they believe contain a mine. Marking flags does not simplify the game as a cell marked with a flag is not validated by the game. A clever player will use the numbers in the cells to find which cells contain mines. However, sometimes it is not possible to derive all the mines from numbers and the user has to guess the next square to uncover.

In most of the implementations of the game the first move is always a safe move. This guarantees that the first box that is uncovered will not contain a mine and will be empty. There are some slightly different versions of the game where the first move can uncover a mine, but it does not make the game more interesting as at the beginning the player has no information to tell which tile is not a mine. In this project, the most common version of Minesweeper will be used which guarantees that the first move always uncovers an empty cell\footnote{correction: it was later found that the first move should be guaranteed to be safe, not zero. It is a small difference that shall be discussed later.}.

\hfill

\begin{tcolorbox}[colback=gray!5!white,colframe=gray!100!black,title=Minesweeper Difficulty levels]
  Typically, Minesweeper has three difficulty levels:
  \begin{itemize}
    \item \textbf{Beginner}: a $\text{9} \times \text{9}$ board with 10 mines. It is a rather easy level and it rarely requires making guesses. The numbers in the tiles are mostly ones, twos and threes.
    \item \textbf{Intermediate}: a $\text{16} \times \text{16}$ board with 40 mines.
    \item \textbf{Expert}: a $\text{16} \times \text{30}$ board with 99 mines. The bigger numbers are more prevalent as compared to other levels but seeing 7 and 8 is still rare.
    
  \end{itemize}
\end{tcolorbox}

Most of the Minesweeper implementations allow the player to set a custom height, width and number of mines in the game. This project will mainly use Beginner, Intermediate and Expert difficulty levels. As a general rule, the difficulty can also be deduced by finding the density of the mines by dividing the number of mines by the number of cells in a board of size $n \times m$.

\hfill

\begin{equation}
\label{eq:eq_mine_density} 
\text{mine\:density} = \frac{\text{Number\:of\:mines}}{\text{n}\times \text{m}}
\end{equation}

\hfill

It is important to mention that the mine density is only an indication of the difficulty level but it does not always reflect that correctly. For example, if two boards of different sizes have the same density, the bigger one is likely to be more difficult than the smaller one. If a board has a high mine density but has only one or two rows or columns it will be pretty straightforward.


\section{Basic Definitions}
This section covers some basic definitions that will be used throughout this report when discussing Minesweeper game.

\hfill

\begin{tbluebox}{\begin{definition}\normalfont Configuration \end{definition}}
A Minesweeper configuration or board is an ${n} \times {m}$ rectangular matrix of tiles containing numbers between 0 and 8, mines and covered cells. A Minesweeper state configuration can also contain flagged tiles.
\end{tbluebox}

\hfill

A configuration will be assumed to be a valid state of the game at any given staged of the game. A solution to the game is a set of moves that lead to uncovering all the tiles except from those containing mines. A solution can also be seen as marking all the mines of on a given configuration. 

\textbf{Tiles}/\textbf{Boxes}/\textbf{Cells} will be used interchangeably to identify a position in the Minesweeper configuration. A cell $x$ in a configuration can be identified by its coordinates (r, c) where r indicates the row index of $x$ and $c$ indicates the column index of $x$. In this project, the indices will start from 0.

A \textbf{covered} or \textbf{unknown} cell is a cell whose content is not known by the user or has remained untouched since the start of the game. An \textbf{uncovered cell}, on the other hand, is a cell that has been revealed. During the game play, an uncovered cell always contains a number between 0 and 8 as uncovering a mine ends the game. 

\hfill

\begin{tbluebox}{\begin{definition}\normalfont Bordering Cells \end{definition}}
A covered cell or tile that has at least one uncovered neighbouring cell is a bordering cell. 
\end{tbluebox}

\hfill

Bordering cells become relevant when playing Minesweeper as these are potentially next cells to be uncovered. Bordering cells can also be seen as covered cells that border with any cell that have been uncovered. Note that bordering cells are not to be confused with cells at borders of the board which are the cells at the top, bottom rows and left and right most columns.

A \textbf{safe} or \textbf{mine-free} tile is a tile about which it can be concluded with certainty that it does not have a mine.

A \textbf{flagged} or \textbf{marked} cell is a covered cell that is identified as containing a mine. Unless specified otherwise, it will be assumed that a flagged cell has been correctly identified as containing a mine. Unless mentioned otherwise, when a covered cell is mentioned, it will be meant that it is not flagged.

\section{Complexity of Minesweeper} 
\subsection{P vs NP}
Before reflecting upon the complexity of Minesweeper, we have to mention the two most discussed classes of problems in complexity theory namely P and NP. 

\hfill

\begin{tbluebox}{P and NP}
\begin{definition}\normalfont 
  \textbf{P}: A language L is in P if and only if there exists a
  polynomial-time algorithm $A$ such that
  \begin{itemize}
    \item if $w \in \text{L}$ then $A(w)$ outputs yes
    \item if $w \notin \text{L}$ then $A(w)$ outputs no
  \end{itemize}
  
\end{definition}

\begin{definition}\normalfont 
    \textbf{NP}: A language L is in NP if and only if there exists a polynomial $p$ and a polynomial-time Turing machine $M$ such that for all $w \in \Sigma^*$, $w \in \text{L}$ if and only if there exists some $u \in \Sigma^{p(|w|)}$ such that $M$ accepts $(w, u)$. $u$ is also called the certificate and $M$ the verifier \cite{dr_dominik_complexity_theory}.

\end{definition}

\end{tbluebox}

\hfill

The formal definitions can also be extended from languages to problems. A problem is in P if there is a polynomial-time algorithm that solves every input of the problem. On the other hand, a problem is in NP when we can define a certificate of polynomial size and a polynomial-time algorithm that verifies the certificate. 

Polynomial-time algorithms are considered to be more efficient as compared to ones with exponential or worse run time. It is believed that problems in P can be solved efficiently or rather scale efficiently. Whereas considering a problem in NP, its solution can be checked efficiently.  It is known that P $\subseteq$ NP because if the solution to a problem can be found in polynomial-time, its solution can also be checked in polynomial-time. The hardest problems NP are known as NP-complete problems. A problem is \textbf{NP-complete} if it is in NP and NP-hard.  Whether there exist polynomial-time algorithm for solving every NP problem is unknown. This is why the famous question P $=$ NP is still open. An interesting fact is that for proving that P $=$ NP holds, it is enough to find a polynomial time algorithm for solving any one of the NP-complete problems as any NP problem can be reduced to an NP-complete problem in polynomial time. Hence, solving one NP-complete problem in polynomial time would mean that any problem in NP can be solved in polynomial time.

For the purposes of this project, it is sufficient to know that problems in P are the ones that can be solved in a reasonable amount of time whereas NP-complete problems are hard and assumed not to be solvable in polynomial time on every instance.

\subsection{Minesweeper is hard}
When trying to find the complexity of solving Minesweeper we have to consider what are the different ways of playing Minesweeper. One of these methods is Minesweeper consistency problem which has been used by Richard Kaye to solve Minesweeper \cite{richard_kaye_np_completeness}.

\hfill

\begin{tbluebox}{\begin{definition}\normalfont Minesweeper consistency problem \end{definition}}
Given a Minesweeper configuration in input with potentially flagged mines, can the mines be placed in such a way that is consistent with the uncovered numbers in the configuration?
\end{tbluebox}

\hfill

Kaye showed that, given a configuration, Minesweeper can be solved using Minesweeper consistency problem as follows:

  \begin{enumerate}
    \item Choose a covered tile $t$ and mark it as a mine obtaining a new         configuration $C_m$. Run the consistency problem on the configuration
    \item if $C_m$ is not consistent then $t$ must not contain a mine hence it is a safe tile.
    \item if $C_m$ is consistent then place 0...8 in $t$ obtaining configurations $C_0$, $C_1$, ... $C_8$.
    \begin{enumerate}
      \item If none of the configurations $C_0$, $C_1$, ... $C_8$ is valid, then $t$ contains a mine.
      \item If at least one of the configurations $C_0$, $C_1$, ... $C_8$ is valid, then there are multiple values that can be place in $t$ hence we cannot make a decision. Choose another square and return to step 1.
    \end{enumerate}
  \end{enumerate}

If we run out of squares, then a guess has to be made.

\hfill

\begin{tredbox}
  \begin{theorem}\normalfont 
    Minesweeper consistency problem is NP-complete.
  \end{theorem}
\end{tredbox}

\hfill

Kaye proved that Minesweeper consistency problem is an NP-complete problem by reducing SAT to Minesweeper consistency. This proof concludes that solving Minesweeper using Minesweeper consistency requires repeatedly solving NP-complete problems. Finally, Kaye concluded that solving Minesweeper is an NP-complete problem.

\hfill

It is important to observe that Kaye's paper proves that solving Minesweeper using Minesweeper consistency is NP-complete. It does not prove that it is NP-complete to solve Minesweeper in general i.e. one can argue that there might be a simpler approach to solving Minesweeper that is not NP-complete. In fact most of the people would not use this method when they are playing Minesweeper game. Given a Minesweeper configuration, a player would rather want to infer at least one covered cell as mine-free or containing a mine. Hence, we consider the Minesweeper inference problem \cite{minesweeper_co_np_completeness}.

\hfill

\begin{tbluebox}{\begin{definition}\normalfont Minesweeper inference problem \end{definition}}
\textbf{Input}: A consistent board configuration, containing some revealed cells, correctly flagged mines and the number of hidden mines k.

\textbf{Question}: Is there at least one covered cell whose content can be inferred as mine or mine-free from the available information?
\end{tbluebox}

\hfill

Minesweeper inference is a more natural way for attempting to solve a Minesweeper game when a player is playing it. The player tries to find a cell that is safe or certainly contains a mine. Applying Minesweeper inference after each move allows us to solve the Minesweeper game.

\hfill

\begin{tredbox}
  \begin{theorem}\normalfont 
    Minesweeper inference problem is co-NP-complete.
  \end{theorem}
\end{tredbox}

\hfill

It has been shown that Minesweeper inference problem is co-NP-complete and hence, unless P $=$ co-NP holds, there does not exist a polynomial-time algorithm that can solve Minesweeper using Minesweeper inference. This means that solving Minesweeper using inference requires repeatedly solving co-NP-complete problems which makes Minesweeper a computationally expensive problem to solve using inference.

Whether we use Minesweeper consistency or inference, solving Minesweeper remains  NP-hard and depending on the approach used, playing it is NP-complete or co-NP-complete.

\section{Approaches to solving Minesweeper}
\subsection{Single Point Strategy (SPS)}

One of the simplest methods of playing Minesweeper is to use this strategy. It makes use of the flagged cells, assuming that they are correctly placed. Single point strategy is called so because, given a cell $x$, it only takes into consideration the immediate neighbours of a cell \cite{kasper_pederson_strategies_for_game_playing}.

\begin{equation}
\label{eq:sp_eq} 
\text{MinesLeft($x$) = Value($x$) - FlaggedNeighbours($x$)}
\end{equation}

\begin{enumerate}
    \item if MinesLeft($x$) = 0 then all of its non flagged neighbours of $x$ are safe and hence they can be cleared.
    \item if MinesLeft($x$) = CoveredNeighbours($x$) then all of the covered neighbours of $x$ are mines.
    \item in all the other cases the content of $x$'s neighbours is unknown. Hence, we pick another tile and return to step 1.

\end{enumerate}

\begin{figure}[ht]
    \centering
    \subfloat[]{%
        \includegraphics[width=0.40\textwidth]{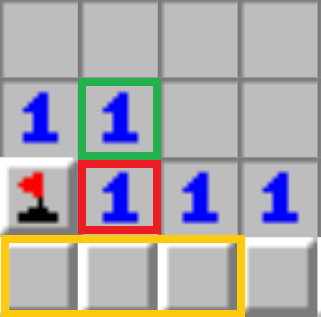}%
        \label{fig:single_point_strategy_1}%
        }%
    \hfill%
    \subfloat[]{%
        \includegraphics[width=0.40\textwidth]{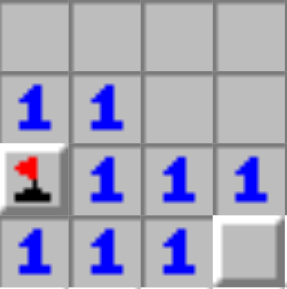}%
        \label{fig:single_point_strategy_2}%
        }%
    \caption{Single Point Strategy}
    \label{fig:single_point_strategy}
\end{figure}

The cell surrounded with the green square in Figure~\ref{fig:single_point_strategy_1} contains one and has exactly one hidden cell as neighbour hence case 2 of SPS is applicable, the only covered neighbouring cell is a mine and has been flagged. In Figure~\ref{fig:single_point_strategy_1}, the cell highlighted with red square is a one and has exactly one flagged neighbour hence all of its covered neighbours can be cleared as shown in Figure~\ref{fig:single_point_strategy_2}.

The single point strategy is useful in solving configurations with low density such as beginner level. But as the mine density increases, this technique starts to become ineffective for intermediate and especially expert level as information of immediate neighbours is not enough.

When playing Minesweeper with SPS, in the first move, a random cell is selected and uncovered without applying single point strategy as all cells are covered and no clues can be found about placement of mines. Let $Q$ be the set of all uncovered cells neighbouring bordering cells. Each $x \in Q$ will contain a number and have at least one covered neighbour. $Q$ is useful to keep track of cells that contain some information which can be used in the subsequent moves. It is important to know that after applying SPS on a cell, if step 1 or 2 apply then all of the covered neighbours of $x$ are either uncovered or flagged. This makes $x \notin Q$ in the next configuration.

\linespread{1.2}{\begin{algorithm}
    \caption{Single Point Strategy}
    \label{algo:single_point_algorithm}
    \begin{algorithmic}[1]
        \Statex
        \State {$Q \gets \{\}$}
        \While {game not lost \textbf{OR} game not won}
            \If {$Q = \{\}$}
                \State $x \gets$ Random(CoveredCells())
            \Else
                \State $x \gets$ First($Q$)
            \EndIf
            \If {MinesLeft($x$) $=$ Count(CoveredNeighbours($x$))}
                \State $\text{toFlag} \gets$ CoveredNeighbours($x$)
                \State FlagCells(toFlag)
                \State $Q \gets Q$ $\cup$ UncoveredNeighbours(toFlag)
            \ElsIf {MinesLeft($x$) = $0$}
                \State safeCells = CoveredNeighbours($x$) 
                \State $\text{newCells} \gets$ Uncover(safeCells)  \Comment returns cells uncovered in this move
                \State $\text{newBorders} \gets$ BorderingCells() $\cap$ CoveredNeighbours(newCells)
                \State $Q \gets Q$ $\cup$ UncoveredNeighbours(uncoveredBorders)
            \EndIf
            \State $Q$.remove($x$)
        \EndWhile
    \end{algorithmic}
\end{algorithm}}
\break
\subsection{Constraint satisfaction problem}
\label{ch:literature_CSP_solver}

A constraint satisfaction problem, or a CSP, is a mathematical representation of a problem consisting of states whose solution/s must satisfy a set of constraints. A constraint satisfaction problem consists of three components: a set of variables, the domains for each variable and a set of constraints. A CSP-solver has the purpose of finding an assignment for every variable such that all the constraints are satisfied. Such an assignment is called consistent \cite{chris_minesweeper_as_a_csp, becerra_algorithmic_approaches_to_playing}.

A Minesweeper configuration can clearly be seen as a constraint satisfaction problem. Each covered cell is represented as a boolean variable $m \in \{0, 1\}$. The value 0 indicates that the square is mine-free and 1 indicates that it contains a mine. When a cell $x$ is uncovered, a new constraint is added: sum of neighbouring variables of $x$ equals Value($x$). A constraint will be represented as a linear equation that sums up to an integer $n \in \{0,...,8\}$. The number $n$ is the label of the constraint. Clearly, this already covers Single Point Strategy as the constraint label 0 would mean that all the covered neighbours of $x$ are safe. If the label equals the number of variables, then all of the covered neighbours of $x$ are mines.

\begin{figure}[ht]
    \centering
    \subfloat[]{%
        \includegraphics[width=0.40\textwidth]{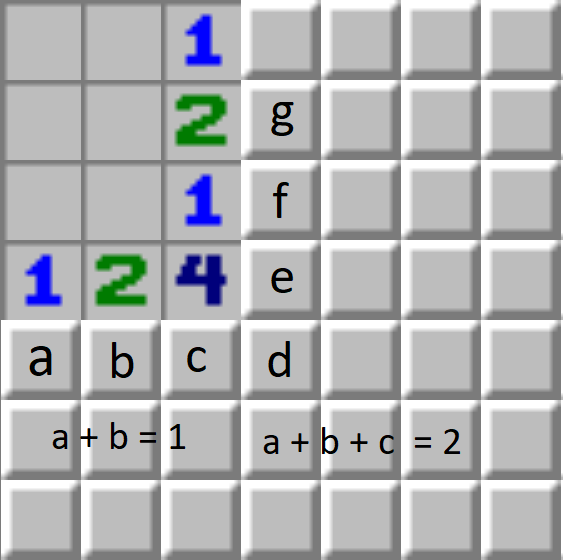}%
        \label{fig:csp_equations}%
        }%
    \hfill%
    \subfloat[]{%
        \includegraphics[width=0.40\textwidth]{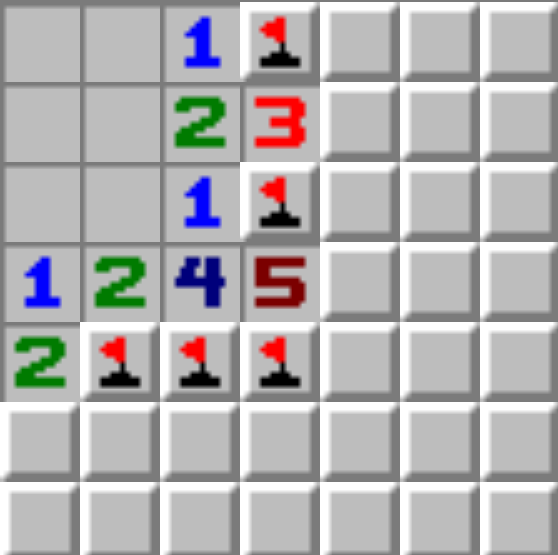}%
        \label{fig:csp_equation_solved}%
        }%
    \caption{Example Adjacency constraint}
    \label{fig:csp_strategy}
\end{figure}

A CSP solver can do more than single point strategy. For example, in the Figure~\ref{fig:csp_equations} $a + b = 1$ must hold because tiles marked $a$ and $b$ are the only covered neighbours of 1. Similarly, $a + b + c = 2$ must hold since $a$, $b$ and $c$ variables neighbour 2. From the equations $a$ + $b$ = $1$ and $a$ + $b$ + $c$ = $2$ it can be concluded that $c$ = $1$ hence $c$ contains a mine. This conclusion cannot be drawn from single point strategy as the information in immediate neighbours is not enough. Likewise, the equations $b$ + $c$ + $d$ + $e$ + $f$ = $1$, from cell at coordinates (3, 2), and $e$ + $f$ + $g$ = $1$ from cell at coordinates (2, 2) allow us to conclude that $b + c + d = 3$ hence $b$, $c$ and $d$ are all mines. Solving these equations and a couple of more uncovers all the bordering cells in configuration from Figure~\ref{fig:csp_equations}.

CSP is the best approach for solving Minesweeper. There are still cases when CSP cannot determine what next move to make and hence a guess has to be made. The success rates of CSP for solving Minesweeper are 91.25\%, 75.94\% and 32.90\% for Begineer, Intermediate and Experts levels respectively \cite{becerra_algorithmic_approaches_to_playing}. Higher success rates can be reached by combining CSP with probabilistic approach that finds the safest cell in case of a forced guess instead of a random one. But we shall stick to success rates achieved with CSP only as we analyse single strategies for playing Minesweeper, not a combination of these. The success rates of CSP-solver are relevant for this project and will become a point of reference to compare with when new learners will be assessed.

\chapter{Methodology}
\section{Game implementation}

As specified in the project objectives the game implementation is the first part of the actual implementation of the Minesweeper learner. A custom implementation of the game or an existing open source implementation were the potential options. A custom game implementation using Java was chosen. The game implementation provides simple API calls for setting up the board and performing different moves on the board. A simple user interface was also created for playing the game and testing that the game is played according to the rules. Playing the games automatically using the learners does not require using the user interface.

\section{Neural networks}
Neural networks are the machine learning approach that was chosen for solving Minesweeper. The following architectures of neural networks were trained for solving Minesweeper.
 
\begin{itemize}
    \item Feed-forward neural networks
    \item Convolutional neural networks
\end{itemize}

For each architecture, different configurations of neural networks were tried by changing number of hidden layers, number of neurons in each layer, number of filters, activation function, learning rates, number of epochs and other parameters. The aim was to systematically find the neural network configuration which has the highest success rates in solving Minesweeper game. It is also important to find the fastest neural network that is good enough to have high success rates in solving Minesweeper.

The input of the neural network is the game board or a part of the game board and the output is the safest cell that can be uncovered for the next move. It is also important to mention that there is always a chance that the neural networks make a mistake. The reason behind this is that the number of different board configurations is huge and a neural network cannot guarantee correct output for every input. For this reason flagging was not allowed because if the learner incorrectly flags a cell, then that cell may never be opened again and the game would get blocked. Hence the learner only finds the next cell to be uncovered. It is also important to guarantee that the learner always outputs a covered cell because otherwise the learner may continue to give an uncovered cell as safe and the game would never end.

After implementing a  neural network, another important component of the implementation was running the experiments and saving the results. This acted as a bridge between the neural network learner and the game interface. Since our objective is to run hundreds of thousands or potentially millions of games on a given learner, this part is very important for saving the results of the learner frequently in a file. After every one thousand or ten thousand games in a series, the results of the series was shown in output and appended in a file so that the performance of the learners can be monitored. The metrics saved after every series of games are as follows:

\begin{itemize}
    \item Total number of games played so far
    \item The number of games won in the series
    \item The number of moves made in the series
    \item Sum of the number of cells to uncover when every game ends
    \item Timestamp on at the end of the last game of the series to find the amount of time taken to play the games
\end{itemize}

\section{Deterministic solver implementation}
As discussed in the literature review, CSP solver is the the deterministic solver with highest success rates. The main aim of our project is to create a learner that can outperform or reach similar success rates to the best deterministic solvers for Minesweeper. For a fair comparison, a deterministic solver implementation is needed . Most of the existing research does not reflect upon the execution time of the deterministic solver. It is also important to consider that for a comprehensive analysis of the learner. Therefore, we need to implement or use an existing deterministic solver to solve Minesweeper for finding its success rates and speed.

\section{Tools}
\label{ch:methodology_tools}
The software and hardware tools used for developing the project are as follows:

\begin{itemize}
    \item Windows desktop with remote access
    \item Java programming language
    \item Eclipse IDE as development environment
    \item Git and GitHub for code versioning
    \item Deeplearning4j library for neural network implementation
    \item Maven for building the Deeplearning4j project and managing dependencies
    \item CSP solver for Minesweeper
\end{itemize}

The remote windows desktop was used for training the neural network without any interruption. The hardware specs of the remote desktop are given as follows:

\begin{table}[h]
\centering
\begin{tabular}{|l|c|}
\hline
Component        &       Description      \\ 
\hline
CPU brand        &         Intel          \\  
CPU name         &    Xeon  E5-1650 v3    \\
CPU speed        &        3.50 GHz        \\
CPU cores        & 6 physical, 12 logical \\
RAM              &        16.0 GB         \\
GPU              &      NVIDIA GT 710     \\
Operating System &        Windows 10      \\
\hline
\end{tabular}
\caption{Hardware specs of the machine used for running the experiments}
\label{table_desktop_specs}
\end{table}

    \chapter{Game implementation}

\section{A custom implementation}
For the implementation of the game there are two options. The first is to download an existing open source implementation and make appropriate changes to its code to integrate it with the learner. The second option is to implement the game from scratch incorporating all the rules in it. There are many game implementations available online such as executable file, game playable on websites using GUI and game implementations with source code. If the first option is chosen, an open source implementation is required because the game source code is needed for making moves on the game board according to the prediction by the learner.

A custom implementation of the game was preferred over reusing an open source one \cite{geeksforgeeks_minesweeper} \cite{nick_js_minesweeper} \cite{paulius_tkinter_minesweeper} due to the following reasons:

\begin{itemize}
    \item the online versions are mostly written in languages that are not preferred
    \item reusing existing code requires time to enquire and understand the code
    \item most of the open source versions included GUI and the GUI implementation needed to be separated from the game logic
    \item it can be hard to understand an existing implementation and  make changes or adjustments to the code written by someone else
    \item the rules of Minesweeper game are simple and easy to implement from scratch
\end{itemize}

The above observations were made after doing research about different game implementations available online. It was decided that a custom implementation would be the best option because it is easy to code, use and modify if necessary. Although a custom implementation was chosen due to the reasons described above, an open source implementation would come with the advantage that a thorough testing of the features would not be required. The custom implementation required proper testing of the rules of the game.

\section{Choosing the programming language}
An important decision for the project was the choice of the programming language. The chosen language would influence the style of implementation (procedural or object oriented), tools used for creating the neural networks and to some extend the speed of the system. The two shortlisted options for this decision were Java and Python. Java has the advantage of being my preferred language whereas Python provides many machine learning libraries which can be very helpful for implementing the learner. 

Java was chosen for development of the project. Java provides a more elegant object orientation as compared to Python. Java is also the language that I am more comfortable with when programming. Although less known for machine learning, Java also has some machine learning libraries. Deeplearning4j is a Java library for running deep learning on Java Virtual Machine (JVM) \cite{deeplearning4j_docs}. The deterministic solvers which can be potentially used for performance comparison have also been implemented using Java. Therefore, for a fair comparison of speed it was reasonable to choose Java.

Java has the advantage of being faster than Python because Java is a compiled language. Java code is first converted into machine code which is then executed on JVM. On the other hand, Python is an interpreted language not compiled but executed by an interpreter. The interpreter takes one section of code at a time, parses it and executes it on the machine. This overhead makes the execution of interpreted languages like Python slower. When comparing the speed difference between Java and Python, Java has been shown to be around 33 times faster than Python for operations involving matrix multiplication \cite{java_vs_python_speed}.

\section{Design}

\subsection{Encoding}

The game board is represented as an $n \times m$ matrix. Each cell in the matrix represents the respective cell on the board. An encoding is needed because all the possible values of a cell must be represented as an integer value. Each label is converted into an integer value because the neural networks accept numeric values in input. The cell values are encoded as follows:

\begin{itemize}
    \item the numbers between $0$ and $8$ are encoded with the same values
    \item a mine is represented with the value $-1$
    \item a covered cell is represented with the value $-2$
\end{itemize}

\subsection{Game Tasks}

All the rules of the game need to be implemented. The system was divided into three components game, learner and results recorder. The game abstracted all the operations related to Minesweeper game. These operations have been implemented to be executable via API calls. The API calls are not dependent on whether the game is being played by the user via GUI or by the AI learner without GUI. The game implements the following tasks: 

\begin{enumerate}
    \item create a hidden board with mines and numbers placed correctly
    \item create the current state board which is initialized to all cells covered
    \item uncover a cell
    \begin{enumerate}
      \item if the cell contains a mine the game ends
      \item if the cell is a 0, display 0 in the cell and recursively uncover the neighbouring cells
      \item if the cell is a number other than 0, show the number in the cell
    \end{enumerate}
\end{enumerate}

For fulfilling point 1, the hidden game board is initialized as soon as the game is created for an $n \times m$ board with $k$ mines. After creating an $n \times m$ matrix, all the cell values are initialized to 0. Then $k$ mines are placed onto the board one by one. When a mine is placed in a cell, the cell value is set to $-1$ and all the eight neighbours of the cell with a numeric value greater than or equal to zero are incremented by one. Once $k$ mines are added into the board, each cell of the state board is covered hence set to $-2$.

After the game boards are initialized, the game can be played. The player can make a move by uncovering a given cell with index $i$, $j$. When a cell is uncovered, it is checked that the indices $i$, $j$ are valid and, if the indices are correct, it is also checked that the cell is not already uncovered. Since the learner is planned not to flag the cells but only uncover them, flagging feature was not implemented in the game.

\subsection{UML and Algorithm}
Before starting coding the algorithm for game tasks it is useful to design the game using UML class diagram. In the UML diagram the class attributes, public and private methods are defined to have a conceptual model of the tasks of the game before implementation.

\begin{figure}[!htb]
    \centering
    \includegraphics[width=0.7\textwidth]{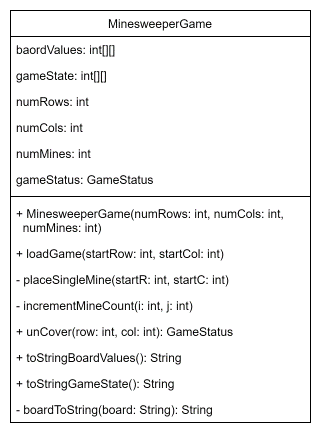}%
    \caption{Game class Diagram}
    \label{fig:uml_game}
\end{figure}

\autoref{fig:uml_game} displays the class diagram of Minesweeper game. The attributes $boardValues$ and $gameState$ respectively refer to the board values and current state board. $numRows$, $numCols$ and $numMines$ contain the size of the board and number of mines. The attribute $gameStatus$ is an enum type for indicating the game status which can be playing, won and lost. The method $loadGame$ utilizes $placeSingleMine$ and $incrementMineCount$ to initialize the $boardValues$ with the cell of first move being safe. The method $unCover$ is used to reveal a given cell and $toString$ methods are used to convert the content of the board into a string so that it can be displayed on screen.

\FloatBarrier

Algorithm \ref{algo:place_single_mine} and \ref{algo:uncover_cell} respectively contain the pseudo code for placing a mine on board and uncovering a given cell. These are the two most important operations in the game. Algorithm \ref{algo:place_single_mine} is called from the method $loadGame$ to place all the mines on the board one by one. The mines are place on the board after the user uncovers the first cell because the first cell is guaranteed to be safe. A random cell on the board is selected for placing a new mine given that the random cell does not already contain a mine and is not the cell of first move.

\linespread{1.2}{\begin{algorithm}
    \caption{placeSingleMine(startR, startC)}
    \label{algo:place_single_mine}
    \begin{algorithmic}[1]
        \Statex
        \State {$\text{mineplaced} \gets false$}
        \While {\textbf{NOT} mineplaced}
            \State {$\text{r, c} \gets$ Random(numRows), Random(numCols)}
            \If {r $ \neq $ startR \textbf{AND} c $ \neq $ startC \textbf{AND}  boardValues[r][c] $ \neq $ -1}
                \State boardValues[r][c] $ \gets -1$
                \State mineplaced $ \gets true$
                \For{ \{i, j\} in Neighbours(r,c)}
                    \If { boardValues[i][j] $ \neq -1$}
                        \State {boardValues[i][j] $ \gets $ boardValues[i][j] + $ 1$}
                    \EndIf
                \EndFor
                
            \EndIf
        \EndWhile
    \end{algorithmic}
\end{algorithm}}
\FloatBarrier

\linespread{1.2}{\begin{algorithm}
    \caption{unCover(row, col)}
    \label{algo:uncover_cell}
    \begin{algorithmic}[1]
        \Statex
        \State {mineplaced $ \gets false$}
        \If {gameState[row][col] $ \neq -2$}
            \State \textbf{return} gameStatus
        \EndIf
        
        \If {boardValues[row][col] $ = -1$}
            \State {gameStatus $ \gets \textbf{Lost}$}
            \State {\textbf{return} gameStatus}
        \EndIf
        
        \State {gameState[row][col] $ \gets $ boardValues[row][col] }
        \State {cellsToUncover $ \gets $ cellsToUncover $- 1$}
        \If {boardValues[row][col] $= 0$}
            \For{ \{i, j\} in Neighbours(row,col)}
                \State {Uncover(i, j)}
            \EndFor
        \EndIf
        \If {cellsToUncover $= 0$}
            \State {gameStatus $ \gets $ \textbf{Won}}
        \EndIf
        \State {\textbf{return} gameStatus}
        
    \end{algorithmic}
\end{algorithm}}
\FloatBarrier

\subsection{Testing}

\begin{figure}[!htb]
    \centering
    \includegraphics[width=0.7\textwidth]{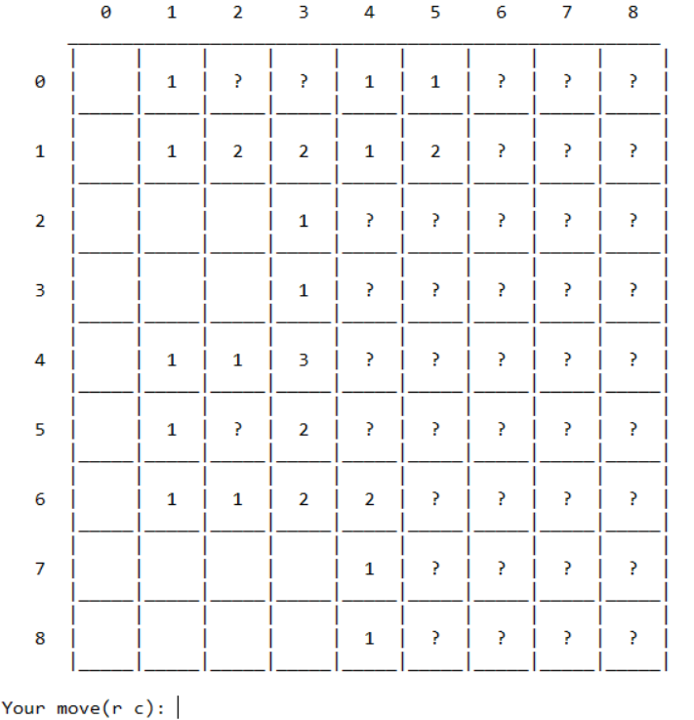}
    \caption{Game play via CLI}
    \label{fig:game_board_string}
\end{figure}

As the game has been developed from scratch, it is essential that the features of the game are properly tested.  Unit testing of the system was done for testing individuals rules of the game. As the system is simple enough and is not expected to have major feature changes in the future, regression testing and other white box testing were not done because these would have cost extra time to the project. Black box testing was conducted which was done by simulating the game play just like a user would play it. For this purpose a command line user interface was developed by converting the board states to string. \autoref{fig:game_board_string} shows the game board displayed on the CLI and moves entered in input for testing the game.

\FloatBarrier

Stress testing of the system is also important to some context. The system should be able to run a large number of games easily without interruption. The system was not overwhelmed when many games were played in succession automatically. The automatic play was done by simply selecting a safe cell at each step and uncovering it. Another aspect of stress testing can be noted by checking the functioning of Algorithm \ref{algo:uncover_cell} for uncovering a cell. When a cell is uncovered and its content is 0 then all of its neighbours are also uncovered by making recursive calls. The question is whether the game would run into stack overflow error if many recursive calls are made. It was tested that for very large boards (such as $1,000 \times 1,000$) having very small mine density (for example less that 0.5\%) the game is likely to run into a stack overflow error. A $1,000 \times 1,000$ board has one million cells and if it has very small density of for instance 0.4\%, the first click would clear almost the entire board and make too many recursive calls to uncover. This error could be fixed by using an iterative approach for $unCover$ method but the games of very large board size are not relevant to the purposes of our project. For this reason it was decided that the recursive approach would be kept as it is simpler, more readable and does not face any errors on the three difficulty modes of Minesweeper.

\section{Building the project}
This section discusses how the project is compiled and built for running the game along with the learner. As mentioned in \autoref{ch:methodology_tools}, the project implementation includes neural networks built using Deeplearning4j library. Before executing the project, it is necessary to build the project following the requirements of Deeplearning4j library. 

\begin{figure}[!htb]
    \centering
    \includegraphics[width=0.4\textwidth]{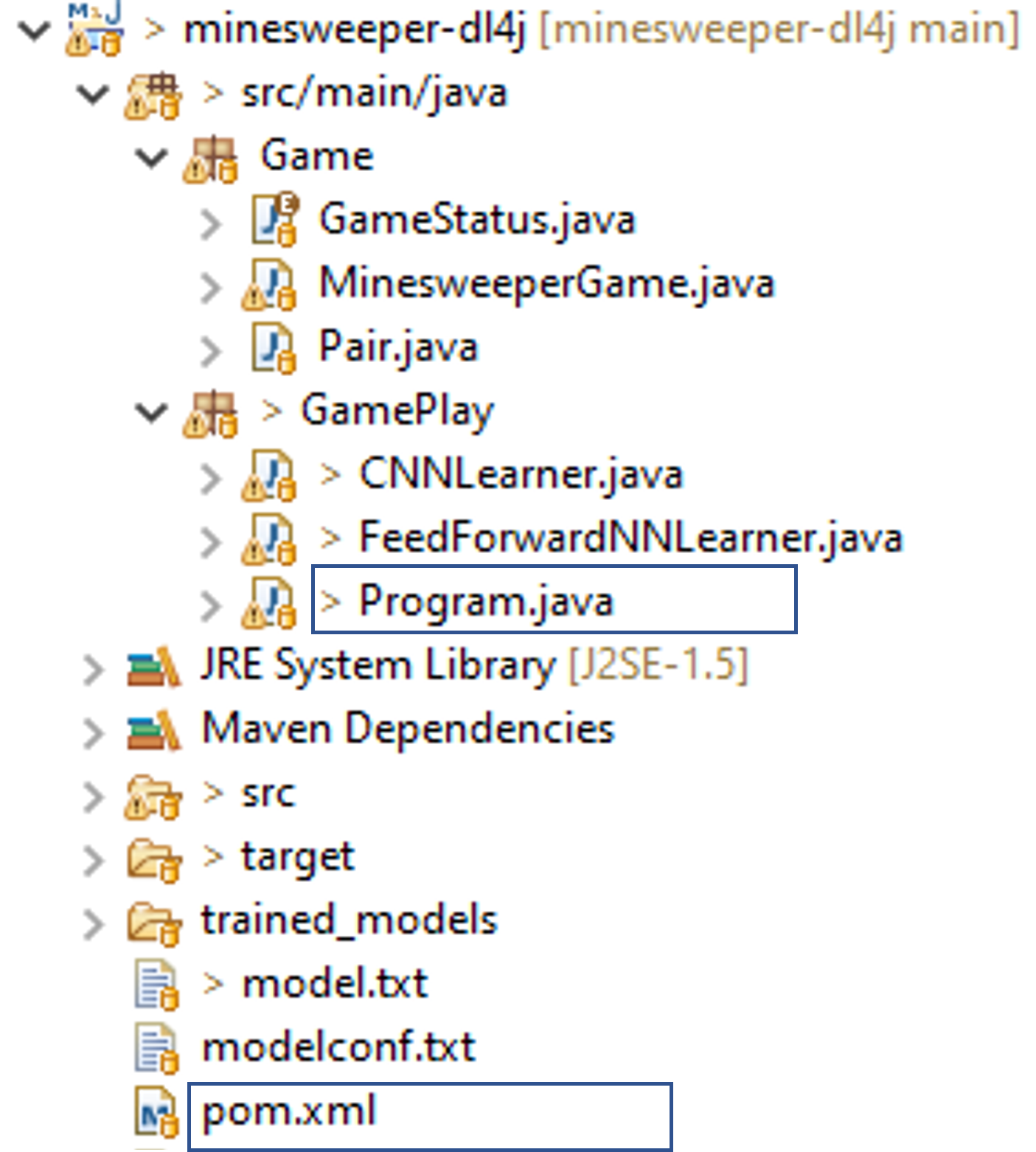}
    \caption{Project structure displayed in Eclipse}
    \label{fig:project_structure}
\end{figure}

Before building the project, we first describe the structure of the project and the most important files. \autoref{fig:project_structure} contains the project structure as displayed in Eclipse IDE. There are two java packages in the folder \texttt{src/main/java}. The package \texttt{Game} contains the classes that implement Minesweeper game. The package \texttt{GamePlay} contains the learners and \texttt{Program} class, point of entry of the project. Another important file is \texttt{pom.xml} file, it contains all the project dependencies from Deeplearning4j library required for building the neural network learners.
It is recommended that any dependencies in the \texttt{pom.xml} file should not be removed, otherwise the neural network learners may not build correctly.
\FloatBarrier

For building the project Maven dependency manager is required. Once the project is opened in Eclipse, building the project is easy. From the Eclipse IDE menu bar choosing \texttt{Project >> Build project} should build the project. Alternatively, clicking on the Run program button should also Build and then run the project. If this does not work further guidance on running the project using Eclipse IDE or IntelliJ IDEA can be found on the Deeplearning4j documentation \cite{deeplearning4j_getting_started}.

\begin{figure}[!htb]
    \centering
    \includegraphics[width=1.0\textwidth]{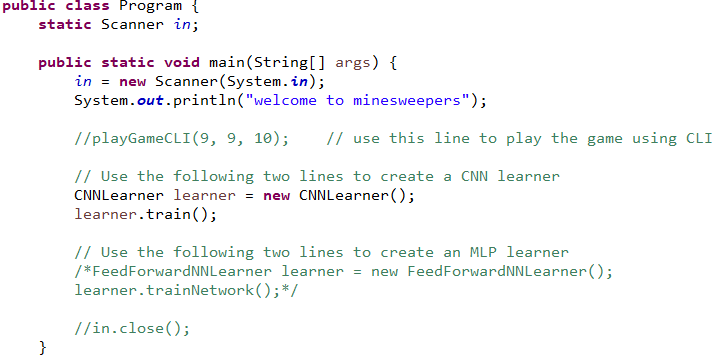}
    \caption{Main method in the Program class where a CLI game play or the learners are instantiated}
    \label{fig:program_class_main_method}
\end{figure}

\autoref{fig:program_class_main_method} displays the \texttt{main} method of the \texttt{Program} class, the point of entry of the project. Once the project is loaded, a default untrained CNN learner will be loaded and shall start training. To play a game via CLI or to train an MLP learner, simply follow the guidelines from the comments. Finally, if an amendment such as loading an existing model is desired, this can be done by following the guidelines in the file of the specific learner type. Further changes or tweaks such as changing the game play of a learner or the configuration of a learner are not recommended unless Deeplearning4j library is studied from its documentation.

\FloatBarrier

    \chapter{Neural Network Learner}
\label{ch:neural_net_learner}
\section{What are neural networks}

\begin{figure}[!htb]
    \centering
    \includegraphics[width=0.7\textwidth]{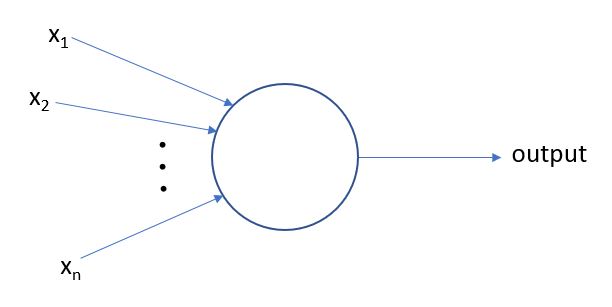}
    \caption{A Neuron}
    \label{fig:artificial_neuron}
\end{figure}

Neural networks are a machine learning technique inspired human brain, mimicking the connections between biological neurons in human brain. This characteristic of the neural networks allows them to reflect aspects of human behaviour for detecting patterns and solving different kinds of problems \cite{ibm_neural_network_intro}. Neural networks are composed of neurons, a neuron has $n$ inputs, $n$ weights, a bias and an output value as shown in \autoref{fig:artificial_neuron}. Multiple layers of such neurons, each layer containing several neurons, are combined to obtain a neural network.

\FloatBarrier

A neural network can also be perceived as a function. Given a dataset comprised of inputs and outputs, the neural network is used to approximate the function that maps the inputs of  the dataset to the outputs. The objective of using this machine learning approach is to find the best approximation function mapping for the given dataset. For this purpose neural networks perform two operations, forward pass and backpropagation. Forward pass is used to obtain the output values for a given input by using weights and biases of each layer. This is done by simple product and addition operations. Then the error value between the output obtained and the correct output is calculated. Finally, the error value is used to perform backpropagation, which adjusts the weights and biases to minimize the error.

\section{Why neural networks}
Many machine learning techniques have been used to solve Minesweeper such as case based reasoning, rule induction \cite{inductive_multirational_learning_minesweeper} and genetic algorithms. According to the indication of the project supervisor these techniques did not reveal to be highly successful in projects from previous years. There is a lack of research in the usage of neural networks for solving Minesweeper. The research paper based on Q-learning for Minesweeper \cite{training_a_minesweeper_solver_q_learning} recommends a neural network learner for solving Minesweeper as future work. A source from Youtube showed that their neural network learner had an impressive success rate of 54\% on Beginner mode of Minesweeper \cite{codinglikemad_minesweeper_neural_network}.

Neural networks were chosen because of the following reasons:
\begin{itemize}
    \item Neural networks are well known for detecting patterns and applying them for classification tasks. Minesweeper has patterns which can be very useful for playing the game \cite{minesweeper_patterns}.
    
    \item Neural networks vary in types, complexity and other parameters which gives a lot of flexibility in trying different configurations and architectures of the neural networks. The success rates of a harder level may be improved by using a more complex neural network as compared to the model for an easier level.
    
    \item The speed of a neural network for a given board size does not vary no matter how hard the game configuration is. Memory based approaches, such as case-based reasoning, get slower as more states are discovered. CSP solver gets slower on harder cases. On the other hand, once a neural network is trained, it will always have the same speed for solving a game, no matter how long it has been trained and how rare or hard the configuration may be.
    
    \item There is a lack of research in the usage of neural networks for solving Minesweeper. If neural networks are shown to be highly successful, this will become the core aspect of this project.
\end{itemize}
 
\section{Inputs and outputs}
\label{ch:neural_network_inputs_outputs}
Whenever a piece of computer program is designed, it is essential to decide the inputs and the outputs. Neural networks are no exception and this decision needs to be taken for neural networks too. It is clear that the input of this learner shall be the game board or some encoding of the game board because the board contains all the information or features that are required to make a prediction about the next move. It is also important to note that depending on the type of neural network a certain formatting of the game board may be required. For example, an MLP only accepts one dimensional inputs whereas the game board is a 2d matrix. On the other hand, a convolutional neural network accepts 2d matrices in input.

\begin{figure}[!htb]
    \centering
    \hspace*{-0.3in}\includegraphics[width=1.1\textwidth]{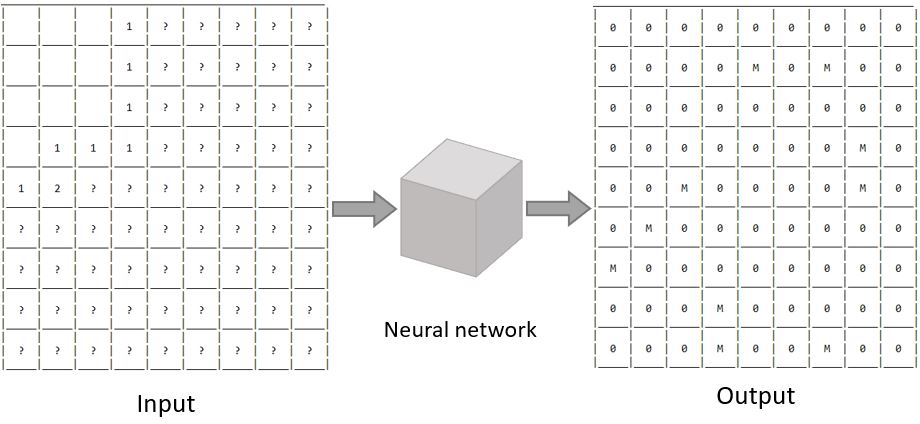}
    \caption{Input and output of the neural network}
    \label{fig:neural_net_io_bin}
\end{figure}

The next step is to decide the outputs of the neural network. When a player is playing the game on a given configuration, they would decide what is the next cell that they would uncover. This is also how the automatic game play works. So on first thought it seems appropriate that the neural network outputs the position of the next cell to be uncovered. But very often on a given configuration, there are multiple safe cells. As stated previously, a neural network should be perceived as a function, having the next move as output would violate the definition of function because multiple safe moves are possible on a given configuration. A function should always have one defined output in the codomain set. Therefore, a more reasonable output is to predict whether or not each covered cell is a mine as shown in \autoref{fig:neural_net_io_bin}.\footnote{The \autoref{fig:neural_net_io_bin} does not indicate that each mine must be correctly identified. For example, non bordering cells do not contain enough information in neighbouring cells and therefore may not be expected to be correctly identified.} Now the next move is merely a piece of information which can be deduced from the output of the neural network. For example, the next move could be a cell among the bordering cells.

\FloatBarrier


\section{Deeplearning4j library}
For building the neural networks there were two options available. The first option was to use a library for building the neural networks and the second option was to code the neural networks from scratch. As the library built neural networks brought many advantages to the project, a library based implementation of neural networks was preferred. Deeplearning4j was the chosen library as it is the java library the provides the deep learning framework. The following reasons summarize why a library was chosen over coding the neural networks from scratch:

\begin{itemize}
    \item Coding the neural networks from scratch is time consuming. As per project objectives, the focus should be on building the best learners in the given time frame rather than creating everything from scratch.
    \item A custom implementation of the neural networks is prone to errors and there is always a risk that bugs are left in the system without ever being discovered which would compromise building the best neural network. The library framework and tools have been already tested and therefore are more trustworthy as opposed to a custom implementation.
    \item The library provides a lot of flexibility because of being in use and hence improved over years. One of the aims is to creates several configurations of neural networks and choose the best ones. The library provides a lot of flexibility in changing the configurations of the neural networks. Deeplearning4j library supports both multi-layer perceptrons and convolutional neural networks learners. On the other hand, a custom framework would required coding both types of networks separately as they differ from each other.
    \item The library based framework allows us to save a model after training and reload it at a later point to resume training or testing. This feature would be very hard to program in a custom framework because saving and reloading all the parameters of a neural networks in a compact manner is a hard task. As it will be discussed, this feature proved to be very useful in many instances.
    \item Deeplearning4j library has been built to create state of the art neural networks that are highly optimized for solving machine learning problems. A custom implementation is very unlikely to have such highly optimized models.
\end{itemize}

Due the above reasons, it is clear that it was in the best interests of the project to use a library based framework for building neural networks. It is important to mention that the library only abstracted the calculations of forward pass and backpropagation that are made by the neural network, all the remaining work still needed to be coded. As it shall be discussed, a significant amount of effort was put into identifying the best configurations of neural network by experimentation.

Many parts of the system were still coded manually. For example, the game implementation, how the learner plays a game needed to be decided and coded. Measuring performance metrics of the learners was also designed and implemented separately. The performance metrics required saving multiple variables in log files, such as the number of games won, the number of moves made, the time taken to play the games, the amount of board cleared and so on. 




\section{Multilayer perceptron}
\begin{figure}[!htb]
    \centering
    \includegraphics[width=0.7\textwidth]{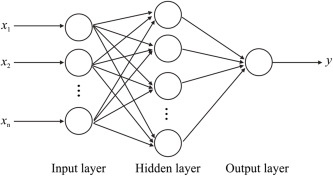}
    \caption{Example of a multilayer perceptron}
    \label{fig:MLP_example}
\end{figure}

Feed forward neural network or multilayer perceptron (MLP) is the most common type of neural network used for classification and regression tasks. This was the first approach used for solving Minesweeper. As shown in \autoref{fig:MLP_example}, a multilayer is composed of multiple layers of neurons. The first layer is the input layer, whereas the last layer is the output layer. The layers between the input and the output layer are the hidden layers. Depending on the complexity of the task one or more hidden layers are used. The task of classification of mines is not expected to be simple, therefore multiple hidden layers were used. A fully connected neural networks was used as each neuron in a layer, except from the first layer, is connected to outputs from all the neurons from the previous layer.

\FloatBarrier

\subsection{Sliding window play}
As discussed in the section \nameref{ch:intro_to_minesweeper}, Minesweeper has different difficulty levels which vary by board size and mine density. The first and most intuitive neural network would take in input the entire $n \times m$ board and output $n \times m$ values indicating whether or not the respective cell is a mine. After examining this approach, it can be said that this is not an efficient and effective neural network for solving Minesweeper. Firstly, the neural network would be very big and would require many nodes in the hidden layers. This makes the neural network notably slow when running making the training and testing very time consuming. Secondly, this type of approach is not scalable because a new network needs to be re-trained from scratch for each difficulty level because every difficulty level has a different board size and the neural network for one board cannot work for another one. Ideally, a good approach should allow the reuse of learning from one difficulty level to another one.

After playing Minesweeper for a few times, it can be noticed that to predict whether or not a cell contains a mine, the most important information is held in the near neighbourhood of the cell. The farther a cell is from the source cell, the less relevant it is for telling whether the source cell has a mine. In the vast majority of the cases the labels in the $5 \times 5$ of a cell are enough. The $5 \times 5$ neighbourhood of a cell is given by the set of tiles taken by moving away by 2 at top, bottom, right and left of the source cell. This was decided in accordance with the guidance from the project supervisor because the projects in previous years showed that a $5 \times 5$ window contains enough information to correctly predict the content of a cell.

\begin{figure}[!htb]
    \centering
    \includegraphics[width=0.7\textwidth]{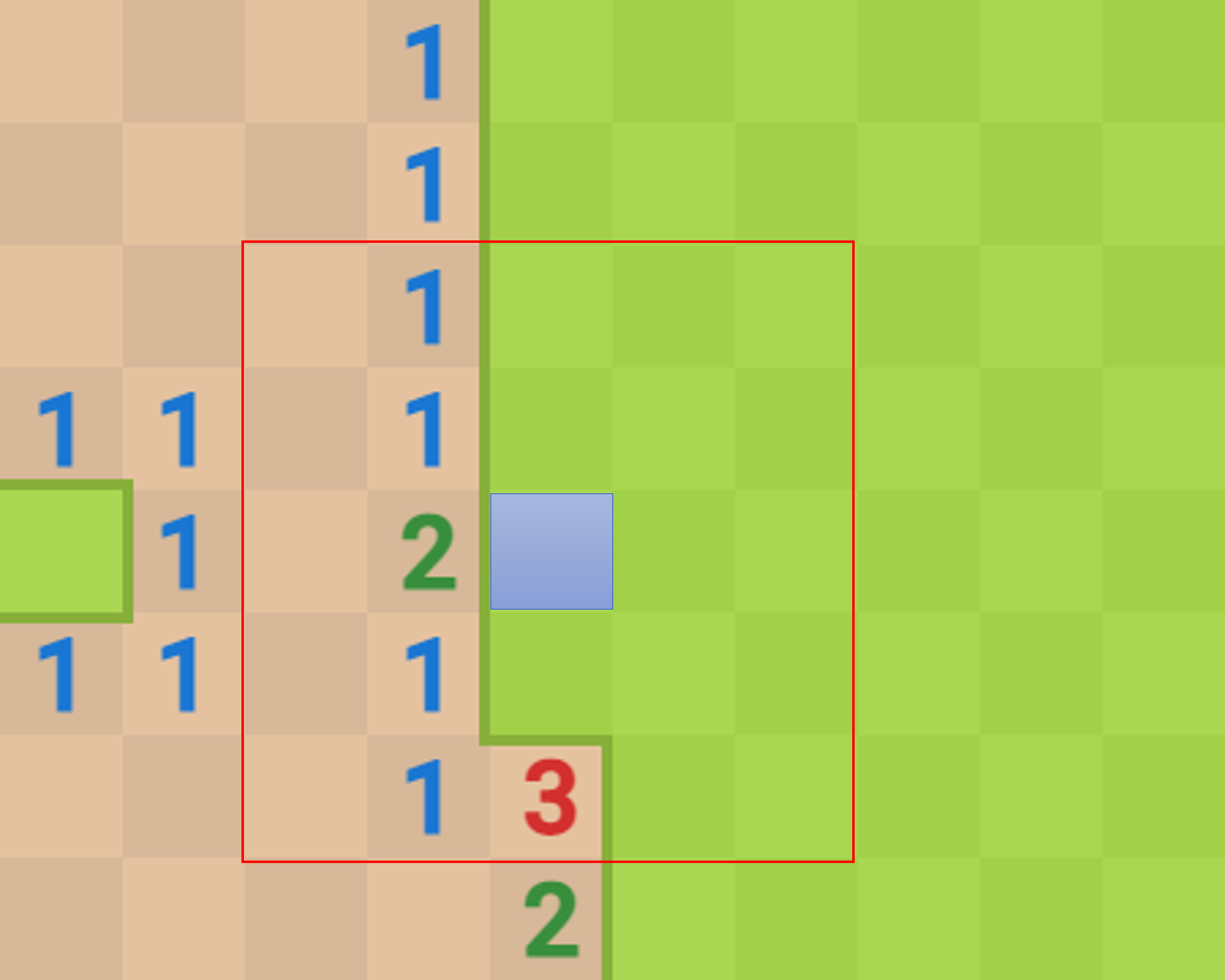}
    \caption{Sliding $5 \times 5$ window}
    \label{fig:minesweeper_5x5_window}
\end{figure}
\FloatBarrier

As shown in \autoref{fig:minesweeper_5x5_window}, to predict the value of the cell highlighted in light blue color, the cells outside the red square can be ignored because these do not provide any information relevant to the highlighted cell. The sliding window becomes the input of the neural network, so the neural network has 24 inputs and predicts the value of the cell in the middle. The 24 inputs are converted into a one dimensional array for being processed by the neural network. The predicted value is a binary indicating whether or not the highlighted cell is a mine. It can also be noticed that the sliding window is very flexible because it does not put restrain on the size of the board. The same neural network can be used for a game of any size or difficulty. On a given configuration, the sliding window is moved on the bordering cells on the board and then a cell classified as safe is chosen for being uncovered.

\subsection{From classification to regression}
As discussed in the section \nameref{ch:neural_network_inputs_outputs}, initially, it was decided that the neural network would be a binary classifier indicating whether or not each cell contains a mine. The output being a binary value soon showed a flaw when the implementation of the network was started. The first issue was that during the training and especially at the start, the neural network is likely to make many mistakes. Sometimes the neural network incorrectly classified that all the cells contain a mine, this consequently caused the game to be blocked because no cell could be uncovered. Another problem was that when multiple cells on borders are classified as safe, it is hard to decide which one to choose for uncovering. There is also a chance that the cell classified as safe is a mistake and therefore when uncovered ends the game. 

Instead of having a binary value as output, it is more useful to have a continuous value between 0 and 1 which indicates the confidence in the prediction. The more the value approaches 1, the more likely it is that the relevant cell contains a mine. On the contrary, if the prediction approaches 0 this means that the cell is more likely to be safe. This type of prediction indicates the amount of confidence in whether or not the cell contains a mine. This solves both of the problems as a move can always be made by choosing the cell predicted with minimum value so the game never gets blocked.  The approach also allows a comparison between the cells likely to be safe and chooses the safest one. 

\vspace{0.2in}
\begin{figure}[!htb]
    \centering
    \includegraphics[width=1.0\textwidth]{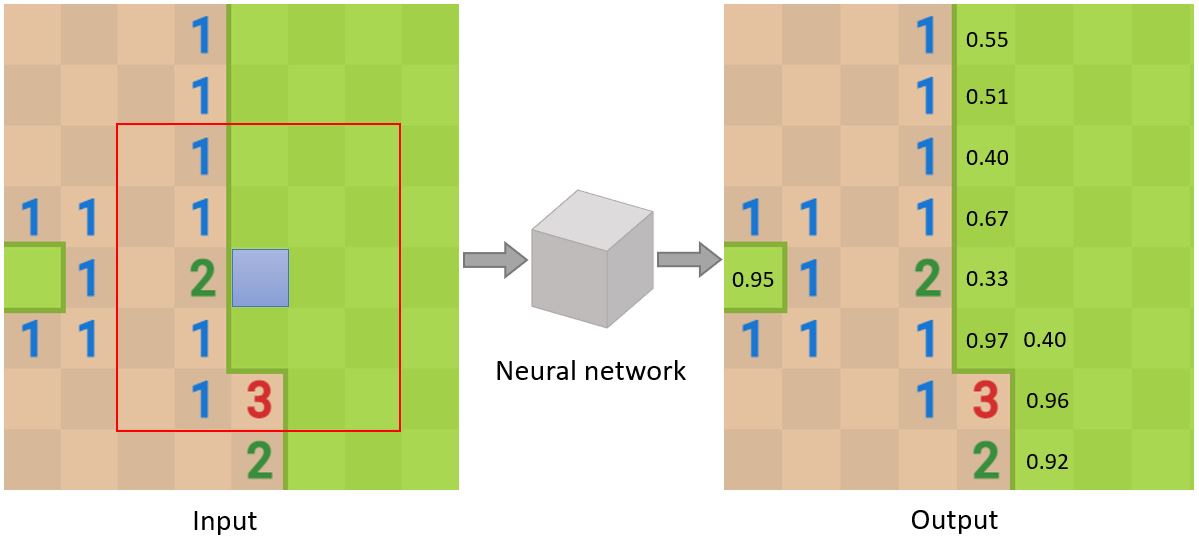}
    \caption{Possible output values of the MLP}
    \label{fig:neural_net_io_reg}
\end{figure}
\FloatBarrier

\autoref{fig:neural_net_io_reg} shows a possible output obtained by moving the window over each of the bordering cells. The movement of the sliding window over the bordering cells is done using a piece of script, whereas the output value of a cell in a given window is calculated by the neural network. In short, the only change made, as compared to the outputs described in section \nameref{ch:neural_network_inputs_outputs}, is that instead of having a binary output it is a continuous output between 0 and 1 which reflects the confidence in the prediction.

\subsection{Parameter tuning}
\label{ch:parameter_tuning}
After defining the type of neural network and the game play using sliding window, the next step is to start training the neural network. For this purpose, it is necessary to define the neural network and its parameters. The parameters of the neural network to be decided are the following:

\begin{itemize}
    \item the number of hidden layers
    
    \item the number of neurons in each hidden layer
    
    \item activation function
    
    \item learning rate
    
    \item error function
\end{itemize}

\subsubsection{Number of hidden layers}
The number of hidden layers and number of neurons in each layer are the most important parameters. These parameters control the topology of the neural network and are decided based on the complexity of the task. According to Jason Brownlee, a well known source for providing tutorials on deep learning, there are no clear guidelines for deciding the number of nodes and layers in a neural network for a given problem \cite{num_of_layers_nodes_mlp_mlmastery}. It is rather a process of discovering the right number of nodes and layers by systematic experimentation. This thought is also shared by many other sources \cite{num_of_layers_nodes_mlp_allaboutcircuits}, some also cited by Brownlee's in his article.

According to the Universal approximation theorem for neural network, one hidden layer is enough for approximating any function. But a single hidden layer is often not enough for complex tasks because it requires too many nodes and the neural network is not practicable anymore. As guided by studying literature, the number of layers and neurons were decided by doing several experiments. For the number of layers it was seen that a neural network with three or four hidden layers started learning the game rules as opposed to one or two hidden layers. 

\subsubsection{Number of neurons}
Once the number of hidden layers was set, the number of nodes in each layer was progressively changed and, after training the network for many games, it the success rates were seen to decide the suitable number of neurons. The following tables shows the number of neurons in four most relevant networks

\begin{table}[h]
\centering
\begin{tabular}{|l|c|c|c|c|}
\hline
Network label  & layer 1 & layer 2 & layer 3 & layer 4    \\
\hline
MLP\_learner1    &  40   &  30   &   20   &  10            \\  
MLP\_learner2    &  50   &  60   &   40   &  20            \\
MLP\_learner3    &  70   &  80   &   60   &  35            \\
MLP\_learner4    &  70   &  100  &   90   &  45            \\  
\hline
\end{tabular}
\caption{MLP learners sizes}
\label{table_mlp_networks}
\end{table}

For each network, the four layers listed in \autoref{table_mlp_networks} are the hidden layers. Each of the listed neural network also had an input layer and an output layer. The input layer had 24 neurons and the output layer had one neuron for all of the networks as the input and output do not change. There were many other learners that were also trained but are not shown in the table because their performance was in similar ranges and did not show any improvements or any deviation from the four learners listed in \autoref{table_mlp_networks}. 

It is important to mention that although the difference in the number of neurons may not seem massive, the number of parameters largely varied between the learners. MLP\_learner1 only had 3,071 parameters whereas MLP\_learner4 had above 22,000 parameters. This shows that increasing the neurons by just 20 or 30 in each layer, increases the number of parameters by a lot. MLPs of smaller sizes than MLP\_learner1 had very weak performances and did not learn the game. Larger neural networks became slower and slower without improving the success rates noticeably.

\subsubsection{Other parameters}
\label{ch:other_parameters}

\textbf{Activation function: } The activation function needs to be chosen for each layer in the neural network. The two main options for the activation functions were the Sigmoid function and ReLU function. The Sigmoid function was very slow in learning the game as compared to ReLU and therefore ReLU function was preferred. For output layer of the neural network, Sigmoid function was chosen because the output of the neural network should be a value between 0 and 1 and only the Sigmoid function produces such values.

\textbf{Learning rate: } The learning rate is usually recommended to be between 0.1 and 0.0001. In our case 0.1 and 0.05 showed the best learning.

\textbf{Error function: } Initially mean squared error (MSE) was used but cross entropy (XENT) was also used. There was no substantial difference between the two in terms of improving the success rates.

\subsection{Training and testing}
Training the neural networks is fundamental for finding a good approximation for solving Minesweeper. Training of different neural networks was used to systematically tune the parameters mentioned in \autoref{ch:parameter_tuning}. Usually, when a learner is trained for a task the training and testing dataset is already given and the learner is trained on the training dataset in iterations. In this project, there is no predefined training dataset. The training dataset is generated as the game is played. When a game is played it is either won or lost. If the game is lost, then all the mines are revealed and if the game is won, then all the mines are correctly identified. In both cases, the bordering cells in different configurations during the game play are the input and the correct output is either a mine or a safe cell which can be found from the final configuration. This is then used to train the neural network via backpropagation.

The neural networks were first trained on the beginner mode. The reason for starting with beginner mode is that the board is small hence playing the games on beginner mode is a lot faster as compared to harder modes. The beginner mode also has easier cases as compared to intermediate and expert mode. Training the neural networks on beginner mode allows us to quickly see whether or not a certain neural network configuration is starting to learn the game. 

In most cases, when a neural network is trained, how long the training should be done is a decision that needs to be taken. It is usually an important decision because training too little does not allow the neural network to learn the task well enough and training too long causes the neural network to overfit the training dataset and be less efficient in solving the task. The overfitting phenomena was surprisingly not seen in the learners for Minesweeper. A convincing reason for this is that there is no predefined training dataset. Usually, there is a predefined training dataset, that has a certain number of samples for example 4,000 samples, keeping the neural network training on the same training dataset causes the model to overfit the training dataset. In case of Minesweeper, there is no predefined training dataset, the neural network is continuously trained on new games that contain several patterns and the patterns or samples seen in training dataset are also the ones seen in testing dataset. So, a potential reason why overfitting was not seen is that the training is not done repeatedly on the same samples as samples keep changing with new games.

After training the learners on beginner mode, the following success rates were obtained:

\begin{table}[h]
\centering
\begin{tabular}{|l|c|}
\hline
Network label    & success rate  \\
\hline
MLP\_learner1    &      24\%     \\  
MLP\_learner2    &     37.5\%    \\
MLP\_learner3    &     40.2\%    \\
MLP\_learner4    &     35.6\%    \\  
\hline
\end{tabular}
\caption{MLP learners trained on beginner mode}
\label{table_mlp_networks_success}
\end{table}

The neural networks had number of layers and neurons as described in \autoref{table_mlp_networks}. The remaining parameters were the same for all the networks and were set as described in \autoref{ch:other_parameters}. As it can be noticed from \autoref{table_mlp_networks_success}, MLP\_learner3 showed the best success rates for solving beginner mode. Unfortunately, even the best neural network so far, had success rates of around 1\% in the intermediate mode and less than 1\% in the expert mode. The project supervisor revealed that the success rate of 40\% on beginner is one of the highest among the learners from previous years therefore, neural network learner has shown an encouraging sign of being effective in solving Minesweeper. Consequently, it was decided that the work on neural network learner would continue and convolutional neural networks were next used for solving Minesweeper.


\section{Convolutional neural network}

After experimenting different multi-layer perceptrons, the success rates of beginner mode showed encouraging signs that a neural network learner could be effective. The next neural network to be used was convolutional neural network. A convolutional neural network (CNN) is distinguished from other neural networks by the usage of convolutional layers. A CNN typically accepts a 2D input and creates a 1D or 2D output. The convolutional neural networks are known for processing images by detecting local features in the image pixels and combining them to make a prediction. 

\begin{figure}[!htb]
    \centering
    \includegraphics[width=1.0\textwidth]{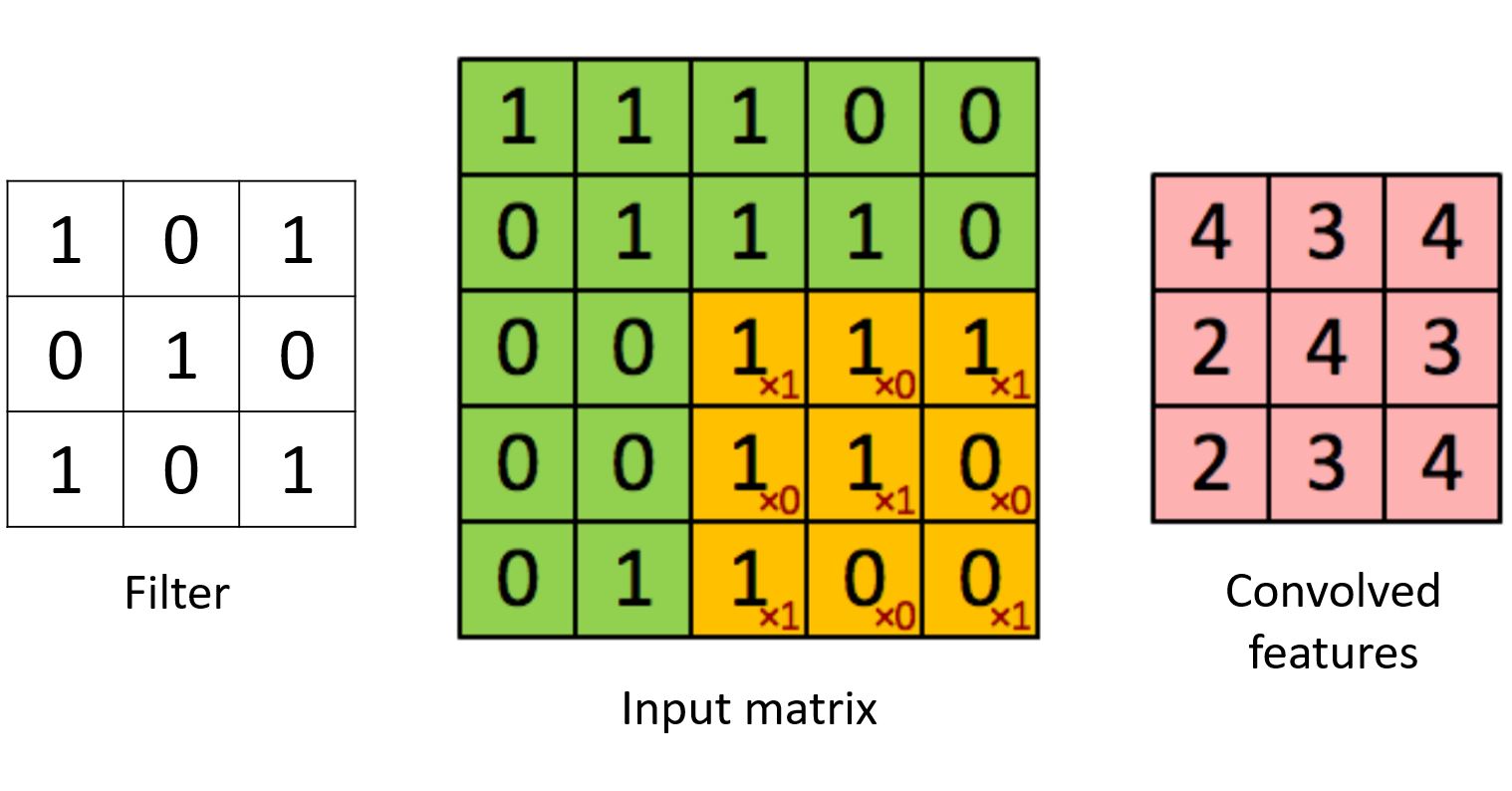}
    \caption[Caption for LOF]{Possible output values of the MLP\footnotemark}
    
    \label{fig:cnn_intro}
\end{figure}

A convolutional layer is composed of a number of filters which are used to perform so called convolution operations. \autoref{fig:cnn_intro} shows a filter and a convolution operation performed on the filter. The filter is moved over the input matrix, after each movement, the values of the filter are multiplied with the respective values of the matrix. The sum of these multiplication is then saved in the output. In \autoref{fig:cnn_intro}, the filter is at the bottom right side of the input matrix for the last calculation and the results of the convolution operation is 4 which is shown in the bottom right cell of the convolved features matrix. Applying the convolution operations in layers allows a learner to detect local features in an image and combine them before finally obtaining a prediction.

\vspace{0.2in}
\FloatBarrier
\footnotetext{Image from https://towardsdatascience.com/a-comprehensive-guide-to-convolutional-neural-networks-the-eli5-way-3bd2b1164a53}

The properties of a CNN learner show that it is a suitable approach for solving Minesweeper because Minesweeper is also played by processing local labels surrounding a covered tile. Furthermore, a research into CSP and Q-learning based solvers recommended using CNN because with this approach local strategies can be generalized on the entire board \cite{training_a_minesweeper_solver_q_learning}.

\subsection{Game play}
Similar to MLP learner, a game play had to be established for CNN learner. Similar to MLP, the output of the CNN learner shall also be between 0 and 1. The aim to create a CNN learner which outputs how likely each tile is to have a mine and then use this information to uncover the safest tile on the board configuration. As discussed earlier, a CNN accepts a 2D matrix in input, thus the board can be given in input without the need of converting it into one dimensional array.

\subsubsection{Padding}

When convolution operations are applied on input matrix, the feature map reduces the size of the board. But the aim of the game play is to get a one to one mapping for each tile of the board configuration, a value between 0 and 1 should be obtained which indicates how safe the cell is. For this purpose, padding of input is supported by convolutional layers to ensure that the output has the same size as the input.

\vspace{0.2in}
\begin{figure}[!htb]
    \centering
    \includegraphics[width=0.5\textwidth]{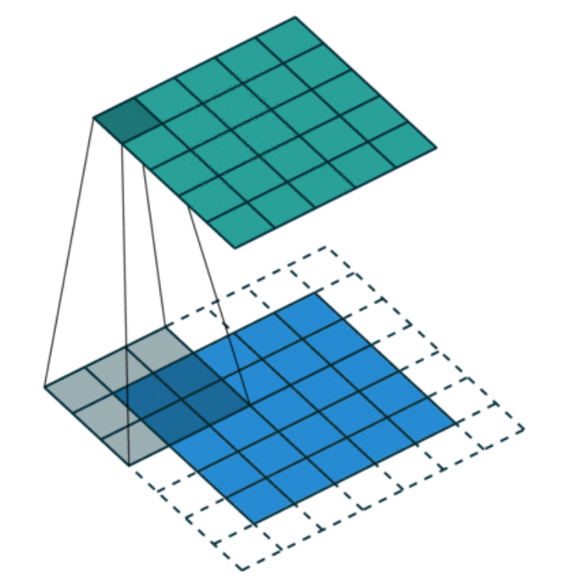}
    \caption[Caption for LOF]{Example of padding on input\footnotemark}
    \label{fig:cnn_padding_example}
\end{figure}
\FloatBarrier
\footnotetext{Image from https://towardsdatascience.com/a-comprehensive-guide-to-convolutional-neural-networks-the-eli5-way-3bd2b1164a53}

\subsubsection{One-hot encoding}
One-hot encoding is an encoding that converts labelled inputs into binary inputs instead of integer encoding. It creates a new binary feature for each label and assigns a value of 1 to the corresponding feature when that label appears. In a convolutional neural network, one input channel is created for each label and the corresponding channel of the label is set to 1 when that label appears in input. For a Minesweeper input board, there are 10 possible input values for each cell, 0 to 8 expressing different numbers and another label for covered cell. So 10 input channels of size of the board are created, when a label 0 appears on position $i, j$ on board, the $i, j$ position of the first channel is set to 1 and position $i, j$ of all the other channels is set to 0.

In machine learning, one-hot encoding is recommended for categorical data, i.e. the data that is labelled \cite{one_hot_encoding_mlp_mlmastery}. There is a strong argument that the inputs of Minesweeper are categorical inputs. The first reason is that a covered cell is a label and not a numeric value. Secondly, the numbers in the uncovered should not be perceived as normal numeric values. It is wrong to assume that a 2 on a cell indicates that the cell next to it is twice as likely to have a mine as opposed to a 1. This is the reason why Minesweeper inputs should be expressed using one-hot encoding for CNN learner.

It was also notice that without one-hot encoding, the CNN learner did not show impressive learning results of Minesweeper. After converting the board into one-hot encoding during pre-processing, the learner showed some improvement in learning the game.



\subsection{CNN configurations and training}
Several CNN configurations were trained to get started and as soon as a configuration showed a good learning of Minesweeper, the parameters of that configuration were noted. After trying a variety of CNN configurations consisting of different filter sizes, number of filters, different activation functions, error functions and learning rates, the following configuration showed success rates of above 40\% on the beginner mode.

\begin{itemize}
    \item layer 1: convolutional layer $20$ $5 \times 5$ filters
    
    \item layer 2: convolutional layer $20$ $5 \times 5$ filters
    
    \item layer 3: convolutional layer $30$ $5 \times 5$ filters
    
    \item output layer: $1$ output between $0$ and $1$
    
    \item sigmoid activation, MSE error function and SGD(stochastic gradient descent) optimizer
\end{itemize}

\begin{figure}[!htb]
    \centering
    \includegraphics[width=0.7\textwidth]{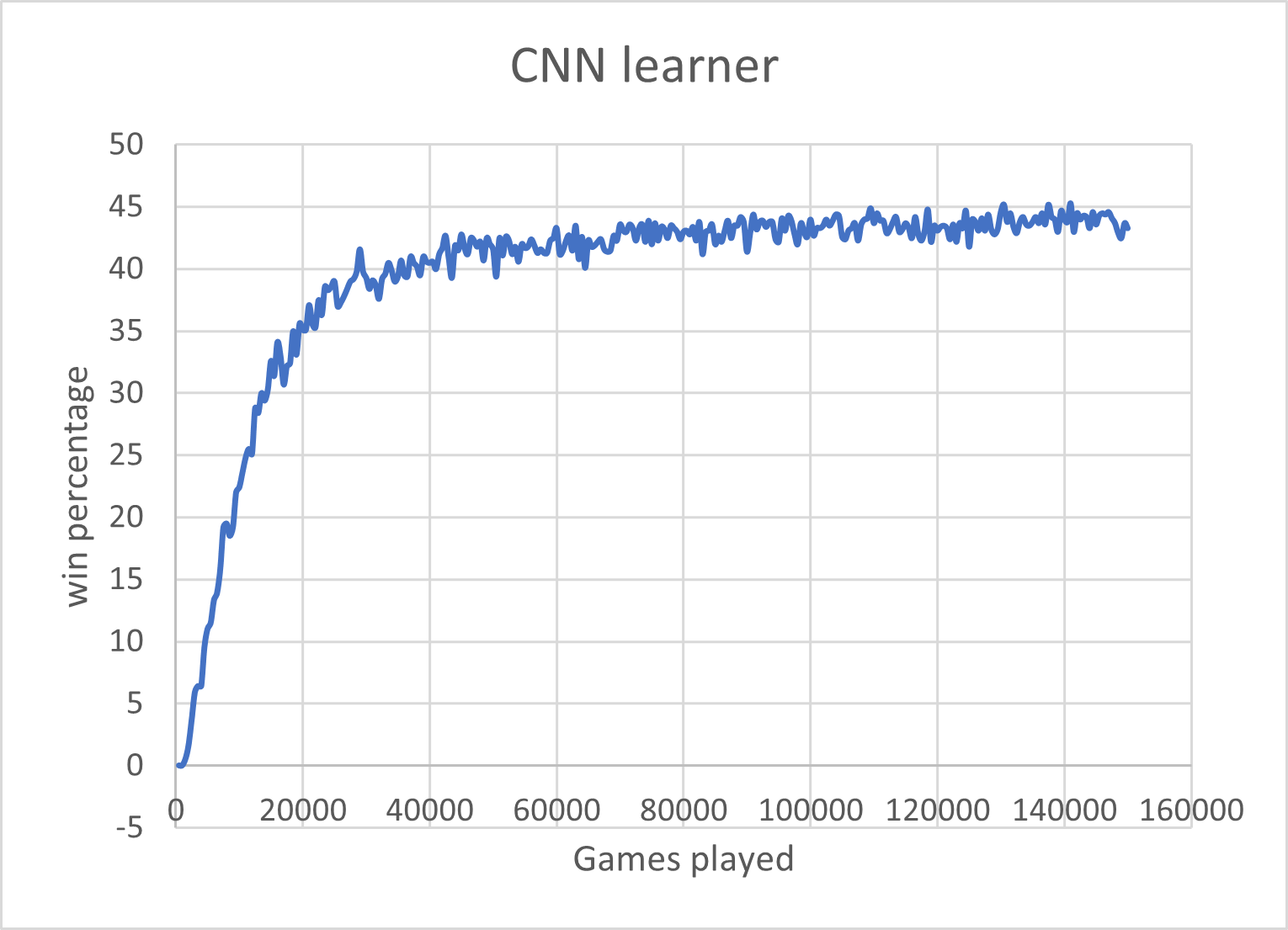}
    \caption{The success rates of the CNN learner during training}
    \label{fig:first_cnnlearner_success_rate}
\end{figure}

The training of this learner is shown in \autoref{fig:first_cnnlearner_success_rate}

\vspace{0.2in}
\FloatBarrier

The learner was trained for about 240,000 games but the convergence of the success rate was witnessed at around 80,000 games. For this reason, the success rates after 150,000 games are not shown in \autoref{fig:first_cnnlearner_success_rate} because they did not contain any improvements.

Further improvement in the success rates was made by increasing the number of filters and changing the activation function to ReLU function. ReLU was used because the use of ReLU function resulted in better performance in MLP learner and the same was witnessed with CNN learner. The learner that significantly increased the success rates on beginner mode described as follows:

\begin{itemize}
    \item layer 1: convolutional layer $25$ $5 \times 5$ filters
    
    \item layer 2: convolutional layer $25$ $5 \times 5$ filters
    
    \item layer 3: convolutional layer $64$ $5 \times 5$ filters
    
    \item ReLU activation, MSE error function and SGD(stochastic gradient descent) optimizer
\end{itemize}

The success rates of this learner on beginner mode was as high as 88\%, very close to the deterministic solver. This high performance encouraged to apply similar configurations for solving intermediate and expert modes of the game. After reaching such high success rates, the intuitive option was to try the same learner for solving harder modes. But using the learner trained for beginner mode to solve harder levels did not show high success rates that were close to deterministic solver. A reason for this is that the intermediate and especially expert mode have a more frequent presence of 4s, 5s and 6s in the labels because the density increases. The learner for beginner mode was not trained on these labels enough to have high success rates on harder modes of Minesweeper. In fact, retraining the same configuration on intermediate and expert level showed better results.

To improve the results further, several learners were created, 24 of them was trained on hundreds of thousands of games and there results were saved in log files. Other learners were not saved because these did not show impressive results.

\begin{table}[h]
\centering
\hspace*{-0.2in}
\begin{tabular}{|l|c|c|c|c|}
\hline
CNN label   &  filters by layer              &  first click  &  game mode &  success rate  \\
\hline
cnnlearner1 & $25$ $5\times5$, $25$ $5\times5$, $64$ $5\times5$ & Zero & beginner     & 88.0\%  \\
cnnlearner3 & $32$ $3\times3$, $32$ $3\times3$, $64$ $3\times3$ & Zero & expert       & 10.1\% \\
cnnlearner4 & $64$ $3\times3$, $32$ $3\times3$, $32$ $1\times1$ & Zero & beginner     & 59.2\% \\
cnnlearner8 & $25$ $5\times5$, $32$ $3\times3$, $32$ $1\times1$ & Zero & intermediate & 62.3\% \\
cnnlearner10& $25$ $5\times5$, $32$ $3\times3$, $64$ $1\times1$ & Zero & beginner     & 91.8\% \\
cnnlearner11& $25$ $5\times5$, $25$ $5\times5$, $64$ $5\times5$ & Zero & intermediate & 72.9\% \\
cnnlearner14& $25$ $5\times5$, $32$ $3\times3$, $64$ $1\times1$ & Safe & beginner     & 84.4\% \\
cnnlearner18& $32$ $3\times3$, $32$ $3\times3$, $32$ $3\times3$, $32$ $3\times3$ & Safe & beginner     & 88.4\% \\
cnnlearner19& $32$ $3\times3$, $32$ $3\times3$, $32$ $3\times3$, $32$ $3\times3$ & Safe & intermediate     & 68.8\% \\
cnnlearner20& $32$ $3\times3$, $32$ $3\times3$, $32$ $3\times3$, $32$ $3\times3$ & Safe & expert     & 21.5\% \\

\hline
\end{tabular}
\caption{Different CNN learner configurations}
\label{table_cnn_learners}
\end{table}

The second column of \autoref{table_cnn_learners} shows the filters in each convolutional layer of the neural network. All the learners had an input layer having 10 channels of size of the board and an output layer with one output value. Some hyper parameters were fixed after experimentation because changing these only worsened the learning of Minesweeper. These parameters are ReLU activation function, MSE error function and updater Adam. Other hyper parameters of the learners can be found in the log file of each learner. Some of the 24 learners have not been shown in \autoref{table_cnn_learners} because these learners do not produce any extra information, the learners displayed in the table already summarize the different configurations.

\subsubsection{A small mistake correction}
After more than half of work had been done on the learner, a small difference in the game version was discovered. The developed game guaranteed that the first click uncovers a 0. This is also the behaviour of the Minesweeper implementation by Google and most of other online sources. However, the project supervisor highlighted that the previous years projects only guaranteed that the first click is a safe cell, namely it is guaranteed not to contain a mine but may contain a number different than 0. After carefully observing the version of Minesweeper solved by David Becerra's CSP solver and other solvers, it was noticed that the first click should be safe and is not guaranteed to be 0. 

For a fair comparison in success rates it was necessary that the first move should be changed to guaranteeing safe cell rather than zero. This required changing the game implementation which was done very quickly as it only required a change of line in the code. This change was expected to reduce the success rates of the learner by a small margin, because when the first click was guaranteed to be 0, more information was revealed. This change in the game was made for cnnlearner14 and all the learners following cnnlearner14.

The column "first click" in \autoref{table_cnn_learners} indicates whether the first click is safe or zero for each learner. When cnnlearner8 was tested on the new version of the game, the success rate dropped to around 75\%. It was clear that the learner needed to be retrained on the game. Retraining was done on learners following cnnlearner11. It can be noticed that for beginner mode, the success rate dropped from almost 92\% to 88.4\% after switching to first click safe.


\subsubsection{Parallel training}
\label{ch:cnn_parallel_training}
Obtaining the best learner required testing many different configurations of neural network. More than 30 different configurations of CNN learners were trained to obtain the 24 best learners, some of them listed in \autoref{table_cnn_learners}. The remaining CNN models were discarded. Training of a CNN  learner often required several days as were trained on millions of games. The training time increased increased even further for intermediate and advanced mode. 

Training all the learners one by one on a single machine would not have allowed to try several configurations and hence would have resulted in a less efficient solver for Minesweeper. It was hence decided that the training would be done in parallel. Two to four learners were trained concurrently, their performance was observed constantly, if a CNN was learning at a very slow pace it was discarded and another learner was trained.

\subsubsection{Slow learning}
For examining whether a CNN model is learning the game, the number of games won were monitored. If the number of games won increased, this was a good indicator that the learner is headed in the right direction. But number of games won does not always increase at a rapid pace. There were many cases when the models were slow at learning the game. In some instances, especially in expert mode, the models took thousands of games before winning the first game. The expert mode on average requires more than one hundred consecutive moves to win a game, that is why a learner requires a considerable amount of training before starting winning. So, if a model was not starting to win any games for thousands of games, it was unclear whether the model is still learning the game or it is failing to learn the game. To observe whether a learner is improving, the number of moves made by the learner was also saved in the log file. If the model improves learning, the number of moves should clearly increase. This showed a lot better whether or not a learner is improving. If the model is learning well, the increase in the number of moves happens much more quickly as compared to winning the games.

    \chapter{Deterministic solver}

The purpose of this project is to compare the performance of the neural network learner with the deterministic solver. After successfully creating some learners for Minesweeper, the deterministic solver is needed for a comparison between the two approaches. As discussed in the literature review, the success rates of the deterministic solvers are already available in different research papers. An implementation of the deterministic solver is needed because the speed of the deterministic solver is the secondary measure of comparison between the deterministic solver and the neural network learner. The deterministic solvers have been discussed in \autoref{ch:lit_rev}. The CSP solver for Minesweeper was chosen because it is the most successful deterministic approach for solving Minesweeper. The functioning of this approach has been describe in \autoref{ch:literature_CSP_solver} and in other sources cited in the literature review \cite{chris_minesweeper_as_a_csp} \cite{becerra_algorithmic_approaches_to_playing} \cite{kasper_pederson_strategies_for_game_playing}.

\section{Custom vs existing implementation}
For using the deterministic solver, there were two options. The first option was to use an existing implementation, ideally a solver quoted in research papers. The second was to implement a deterministic solver from scratch. The second option had to be ruled out because the time was prioritized to be spent on neural network learner. Another reason for not opting for a custom implementation was that the existing approaches have been already tested and well optimized for solving Minesweeper. A custom implementation brings the risk that it may not be as efficient and successful as it should be in solving Minesweeper, thus the comparison would be biased towards the learner. Hence, for these reasons an existing solver was preferred. 
\section{Existing solvers}
\subsection{TeddySweeper}
\begin{figure}[!htb]
    \centering
    \includegraphics[width=1.0\textwidth]{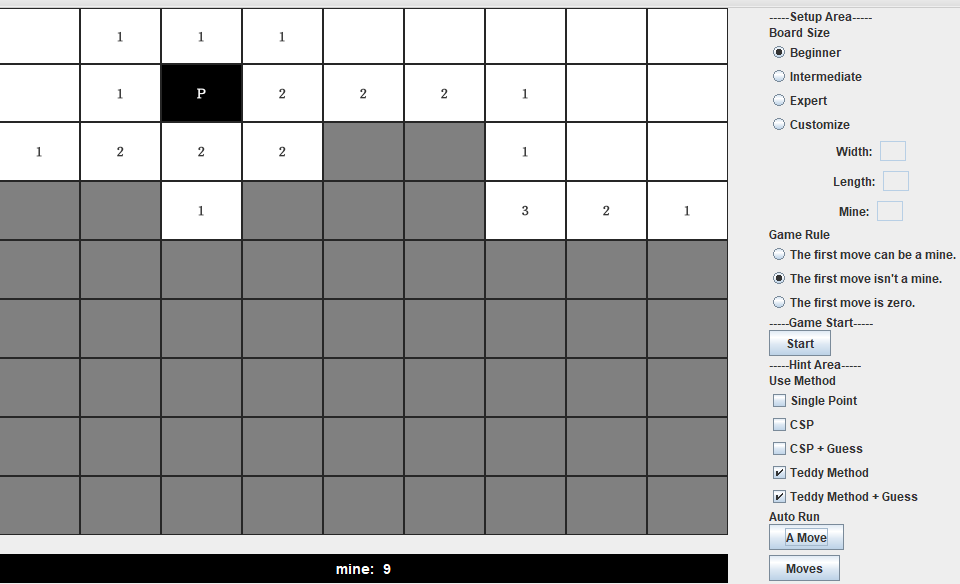}
    \caption{The success rates of the CNN learner during training}
    \label{fig:teddysweeper_gui}
\end{figure}

The first available deterministic solver was TeddySweeper\cite{teddysweeper_solver}. The solver is not available online, the authors of TeddySweeper were requested to provide a copy of TeddySweeper. Tai-Yen Wu, one of the authors, shared a runnable jar file for executing TeddySweeper. As shown in \autoref{fig:teddysweeper_gui}, the executable file is a GUI implementation. For creating a game, the start button has to be pressed manually and the button moves has to be pressed for automatically playing one game. Due the the manual requirement of pressing the button, it was not feasible to use the GUI implementation for running thousands of games. The TeddySweeper project team was requested to provide the source code of the project so that changes can be made to automatically run a number of games. Unfortunately, one of the team members did not consent with sharing the source code and TeddySweeper implementation could not be used for comparison with neural network.

\vspace{0.2in}
\FloatBarrier

\subsection{Becerra's CSP solver}
David Becerra, whose CSP solver has been cited in literature review \cite{becerra_algorithmic_approaches_to_playing}, was also requested to provide the implementation of his CSP solver. Becerra shared the CSP solver and its source code. The CSP solver was built over the Programmer's Minesweeper (PGMS) project. The CSP solver programmed in java was easy to understand as PGMS has an appropriate documentation. This CSP solver was chosen in light of its high success rates which are similar to the other best deterministic solvers for Minesweeper. As described in \autoref{ch:literature_CSP_solver}, it has win rates of 91.25\%,  75.94\% and 32.90\% on Beginner, Intermediate and Expert modes respectively. As required for the project, this solver allows to automatically run many games of a given difficulty mode. After successfully running CSP solver on Minesweeper, it was decided that this solver would be compared with the performance of neural network learner.

\chapter{Evaluation}
\label{ch:evaluation}
\vspace*{-0.1in}
This chapter discusses the results obtained after running the experiments on different neural networks and the deterministic solver. As described in \autoref{sec:intro_aims_obj}, the main objective of this section is to evaluate the success rates of the learners and compare it with that of the deterministic solver. The hardware specifications of the device used for running the experiments have been listed in \autoref{table_desktop_specs}

\section{Performance comparison}
The learners compared in this section are described in \autoref{ch:neural_net_learner}. The labels given to these learners in \autoref{ch:neural_net_learner} will be used for identifying each learner.

\subsection{Convergence}
\label{ch:performance_convergence}
\begin{figure}[!htb]
    \centering
    \hspace*{-0.1in}\includegraphics[width=1.05\textwidth]{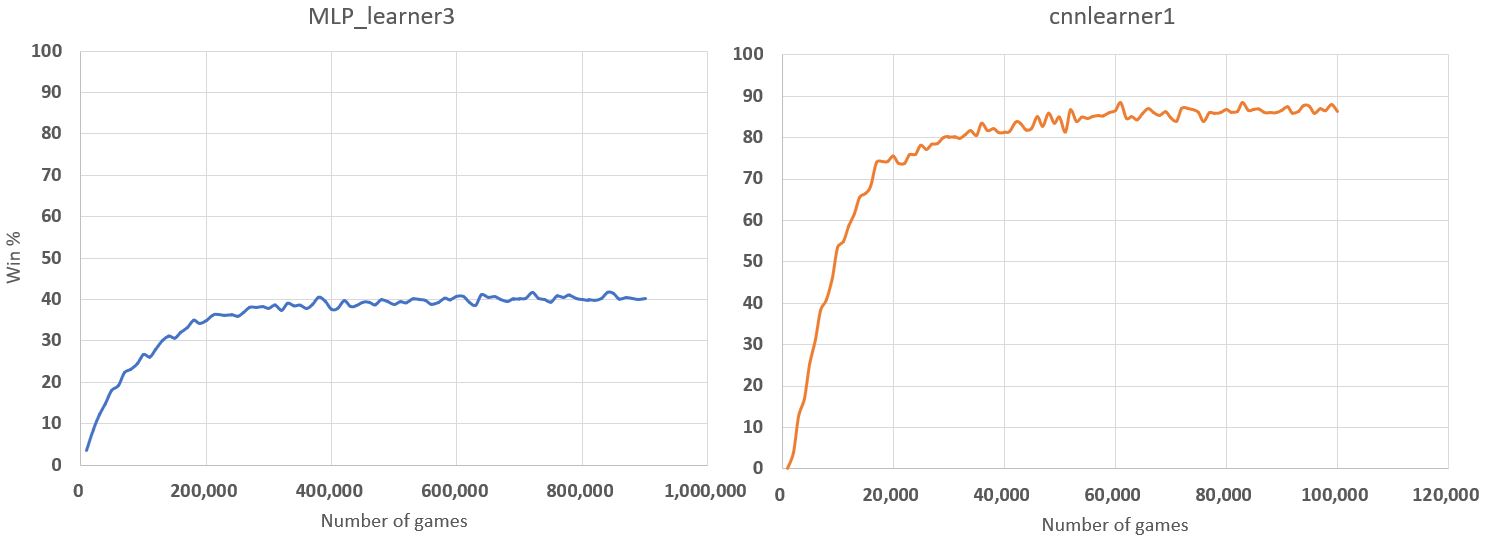}
    \caption{MLP vs CNN training on beginner mode (first click zero)}
    \label{fig:mlp3vscnn1_convergence}
\end{figure}

When a machine learning model is trained on a task, the convergence of the model indicates how long the model needed to be trained before making accurate predictions. \autoref{fig:mlp3vscnn1_convergence} indicates the percentage of beginner mode games won by an MLP and CNN models during training.

\FloatBarrier

As shown in \autoref{fig:mlp3vscnn1_convergence}, cnnlearner1 converged to highest success rates after being trained on around 60,000 games. On the other hand, MLP\_learner3 converged after almost 600,000 games without reaching success rates as high as CNN learner did. Other CNN learners also generally showed to be faster than MLP learners. This clearly indicates that training took a lot more time on MLP learners as compared to CNN learners. It is also one of the reasons why more CNN model were trained as opposed to MLP model.

\begin{figure}[!htb]
    \centering
    \includegraphics[width=1.0\textwidth]{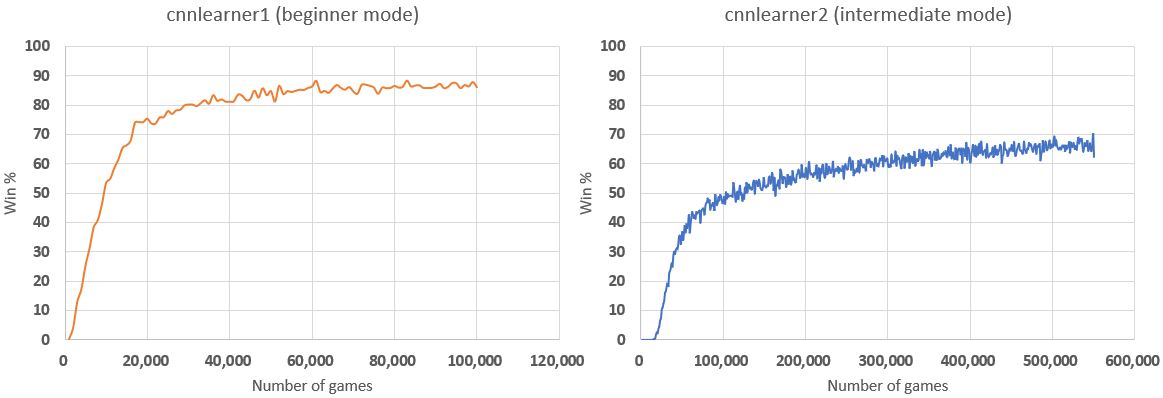}
    \caption{Beginner vs intermediate training}
    \label{fig:cnn1vscnn2_convergence}
\end{figure}

The convergence of a model also depends on the game mode it is trained on. \autoref{fig:cnn1vscnn2_convergence} shows intermediate mode and beginner mode trained on the same configuration of CNN model.
\FloatBarrier

It can be noticed that on the intermediate mode, the model took almost 500,000 games to reach the peak performance whereas on beginner mode the same configuration took only 60,000 to 80,000 games. This shows that training on harder game modes was a lot more time consuming. Furthermore, playing a single game of intermediate mode was a considerably slower than playing a beginner mode game, because the intermediate mode has 256 tiles whereas beginner modes has only 81 tiles. The training got even slower on the expert mode. This resulted in the fact that an extensive training of several models on harder levels could not be done because it was very time consuming task.

\subsection{Success rates}
The success rate is the main criteria of comparison between any two solvers for minesweeper. This section compares the win rates of the learners with those of the CSP solvers.

\begin{figure}[!htb]
    \centering
    \includegraphics[width=1.0\textwidth]{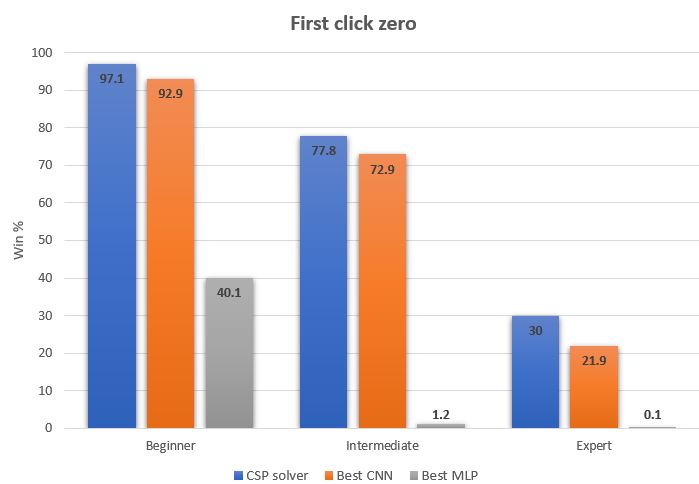}
    \caption{Win rates for optimal CSP solver in the literature and the best CNN, MLP learners}
    \label{fig:firstclick0_comparison}
\end{figure}
\FloatBarrier

Success rates for CSP solver with first click zero are hard to find as most of the CSP solvers are applied on first click safe version of minesweeper. The win rates of CSP solver in \autoref{fig:firstclick0_comparison} are quoted from Kasper Pederson's solver \cite{kasper_pederson_strategies_for_game_playing}. \autoref{fig:firstclick0_comparison} does not reflect the comparison between final models as comparison shall be on the game version with first click safe. From \autoref{fig:firstclick0_comparison}, it is clear that MLP learner is not good enough to compete with CNN learner or CSP solver. In fact, the performance of the best MLP learner is barely visible in intermediate and expert modes. The CNN learner had high success rates however, it always fell short of the CSP solver by a few percentage points.

\begin{figure}[!htb]
    \centering
    \includegraphics[width=0.8\textwidth]{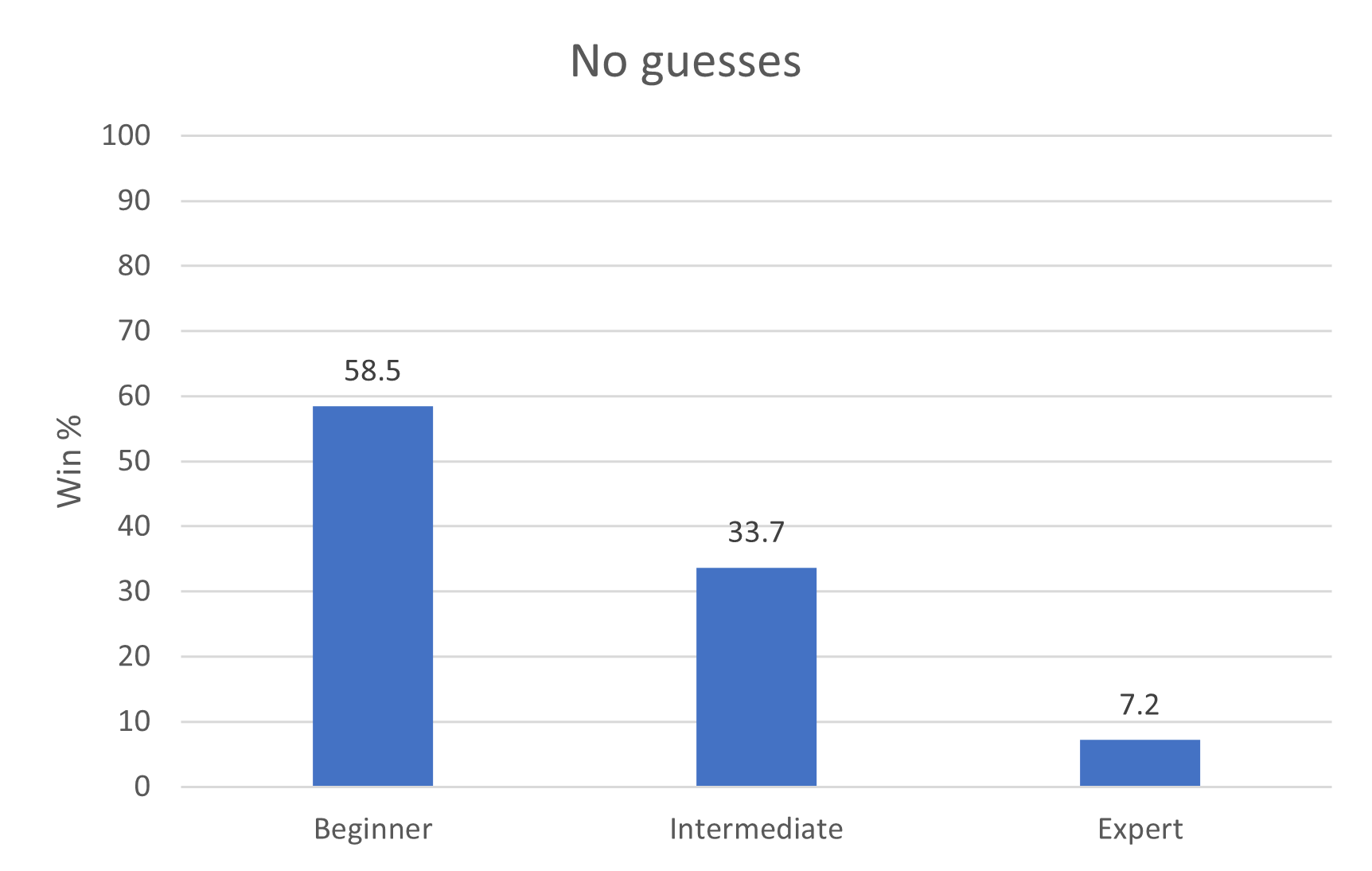}
    \caption{Win rates using CNN learner when guesses are not allowed}
    \label{fig:cnnlearner_no_guesses}
\end{figure}

Guessing usually plays an important role in reaching the high success rates when solving minesweeper. \autoref{fig:cnnlearner_no_guesses} shows the performance of the CNN learner when guessing was not allowed. It can be observed that removing the option of guessing drastically reduced the success rates. The harder the game is the more impact guessing made on the success rates. The intermediate mode had less than half of success rates as opposed to when guessing was allowed. The expert mode was the worst impacted, it approximately had the success rates decreased by a threefold. This shows that, as the game gets difficult, the CNN model makes relies more and more on guessing.
\FloatBarrier

\begin{figure}[!htb]
    \centering
    \includegraphics[width=0.8\textwidth]{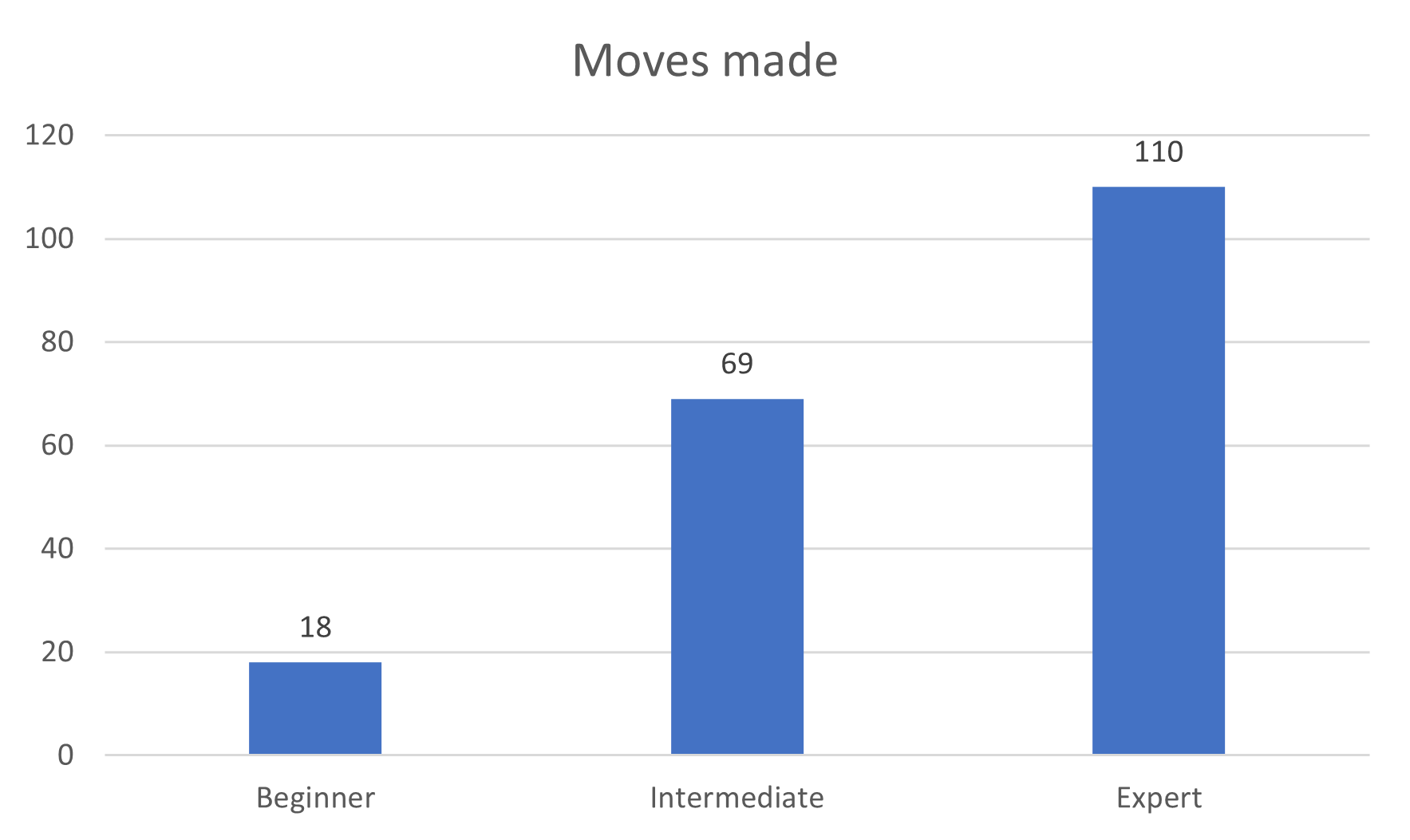}
    \caption{Number of moves made by CNN solver on one game at peak performance}
    \label{fig:cnnlearner_moves_made}
\end{figure}

An interesting finding was the number of moves made by the learner to reach the highest success rates. As the size of board in expert level is larger, it is expected that more moves are made in this level than any other level. \autoref{fig:cnnlearner_moves_made} shows the average numbers of moves made by the best CNN solvers on each game mode. From \autoref{fig:cnnlearner_moves_made} it can be seen that the number of moves made is proportional to the size of the board. The expert board is almost 6 times larger than the beginner mode and the number of moves also has a similar proportion. Similarly, the size of intermediate board is almost 3 times that of the beginner and the same proportion is seen in the number of moves.

\FloatBarrier

\subsubsection{Best model comparison}

We now discuss the performance of the best neural network with the CSP solver. This is the comparison that evaluates the primary objective of the project. For this comparison, the best neural network learner is compared with the best CSP solver. 

\begin{figure}[!htb]
    \centering
    \includegraphics[width=1.0\textwidth]{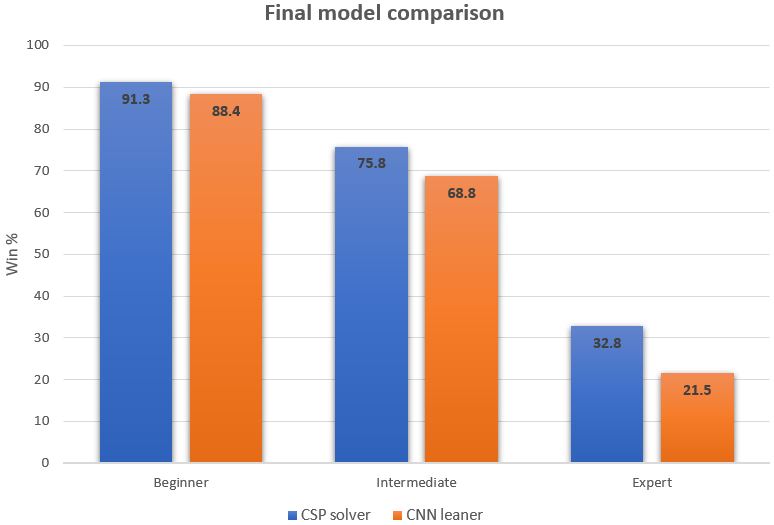}
    \caption{Win rates for Becerra's CSP solver and the best CNN learners for each mode}
    \label{fig:final_comparison}
\end{figure}
\FloatBarrier

\autoref{fig:final_comparison} shows the win rates of the best neural network and the CSP solver. The CSP solver, shared by David Becerra along with the source code, was used for running the experiments. 100,000 games of each difficulty mode were solved using CSP solver to obtain the success rates. Similarly, at least 100,000 games of each difficulty level were played using CNN learners to obtain the respective win rates.

It can be noted from \autoref{fig:final_comparison} that the neural network performs almost as well as the CSP solver, especially in beginner and intermediate mode. Although the neural network has shown to be a competitive solver for minesweeper, it always fell short of the win rates of the CSP solver. The highest difference in performance is witnessed in the expert level. A potential reason for this difference can be deduced from the observations made in \autoref{ch:performance_convergence}. As observed in \autoref{ch:performance_convergence}, reaching the peak performance in harder levels, especially expert mode, proved to require a large amount of training. Although the learner was trained for more than one million games to obtain the success rate of 21.5\%, perhaps further training could have improved the success rates. More reasons for the limitation of the success rates shall be discussed in \autoref{ch:evaluation_limitation_of_learners}.

\subsection{Speed comparison}
As mentioned in the project objectives, the secondary measure of comparison between the learner and the deterministic solver is the speed of the solver. That is how long does a solver take to solve a game. This was not the central question of the research but remains an interesting point to discuss as in some applications the accuracy/speed trade-off can be a decisive factor in choosing one software over the other.

As described in \autoref{ch:cnn_parallel_training}, training of the neural networks was done in parallel for saving time as dozens of neural networks needed to be trained to choose the best among them. It was decided that testing would be done in series mainly because testing was done only on the most successful learners. Moreover, testing multiple of learners in parallel showed to be slower than running in series because, in concurrency, the resources are divided between these learners. Deeplearning4j library allows a learner to run on multiple cores to obtain optimal speed performance. When running multiple learner in parallel, the scheduler decides how long each learner would be executed and thus dividing the resources among the learners. As there was no need to test many learners and parallel execution slowed the performance, it was decided that the best learners would be tested one by one on each mode of the game, to avoid negative effects the speed measures.

Speed comparison was the performance metrics for which experiments needed to be run separately on the machine because information about the speed of the deterministic solvers is not available in the literature. For this purpose, David Becerra's CSP solver, that was used for win rate comparison in \autoref{fig:final_comparison}, was also used for speed comparison. In the test cases done for win rate comparison, the time required for playing the games was also recorded. The time for playing the games was recorded for the neural network learners as well. \autoref{table_speed_comparison} provides the average time taken by the CSP solver and neural network for solving a minesweeper game. The time was averaged from 100,000 games of each difficulty mode.

\begin{table}[h]
\centering
\begin{tabular}{|l|c|c|}
\hline
Mode          &  CSP solver  &  CNN learner  \\
\hline
Beginner      &  $5.07 \times 10^{-5}$ s & $0.053$ s \\
Intermediate  &  $1.23 \times 10^{-4}$ s & $0.28$ s  \\
Expert        &  $6.78 \times 10^{-4}$ s & $1.26$ s \\

\hline
\end{tabular}
\caption{CSP vs CNN solver speed}
\label{table_speed_comparison}
\end{table}

From \autoref{table_speed_comparison} it is clear that the CSP solver largely outperforms the CNN learner in speed comparison. It is important to mention that the speed of the learner is the speed for solving a game during testing. During training the learner was even slower on playing a game because training also include the slow step of backpropagation where the neural network adjusts its weights and biases according to the error function.

Intuitively one may think that the neural network should be faster because it performs simple multiplication operations in forward pass. Now let us try to understand why the speed of the neural networks are as indicated in \autoref{table_speed_comparison}. The CNN models that had highest success rates for beginner, intermediate and expert mode were respectively cnnlearner18, cnnlearner19 and cnnlearner20. Each of these learners had 32,833 parameters. Consider the beginner mode game, any configuration in the board has 81 cells and roughly 18 moves are needed to solve the game. For one move the learner makes 81 predictions indicating chance of each cell having a mine, for 18 moves the learner makes $81 \times 18$ predictions. This follow that for winning one game the learner makes the following number of calculations

\[32,833 \times 81 \times 18 = 47,870,514\]
This means that the learner requires above 47 million calculations for completing one beginner mode game. Similar calculations for intermediate and expert mode show that these respectively require above 579 million ($32,833 \times 256 \times 69$) and above 1.7 billion ($32,833 \times 480 \times 110$) calculations. These numbers do justify the speed of the CNN learner. The number of parameters of the neural network weighed the most in the speed of the learner. The only way of improving the speed was to considerably reduce the number of parameters. Unfortunately, reducing the number of parameters decreased the win rate of the learner which would compromise the main objective of the project. Hence in short the CSP solver largely beats the neural network in speed.

\section{Limitations of learners}
\label{ch:evaluation_limitation_of_learners}
It is important to discuss why, despite reaching close to the performance of CSP solvers, the neural network learners could not beat them. After thoroughly applying several configurations of MLP and CNN learners, it is necessary to mention that the neural networks also have limitations and these could the potential reasons why further improvements in success rates were not seen.

Whenever a neural network is trained on a non trivial task it is never expected to have an accuracy of 100\% on the task because neural networks are approximation functions not exact solvers. In our case, the highest accuracy is the one obtained by the best deterministic solvers, namely the CSP approach. When facing complex tasks, it is usually taken for granted that the neural networks would identify some outliers in the dataset and these are permitted to be misclassified. It is quite possible that in such a large search space like minesweeper, the neural network learner identified some outliers and misclassified them which resulted in lower success rates.

Training of the neural networks took a long time as observed in \autoref{ch:performance_convergence}. It was experimented that in some instances, after seeing the convergence, training the learners further increased the success rates at a slow pace for instance by 0.1\% every 100,000 games. The training had to be stopped at some point because of time limitations. Perhaps, training the learner further could improve the win rates.

It has been discussed in \autoref{ch:neural_net_learner} that the parameters of the neural networks are set by systematically experimenting different configurations of neural networks and choosing the best among the experiments. When a neural network configuration reaches high success rates after days of training, then further improving it becomes very hard because changing one parameter and training the network requires us to wait for days before seeing the effects of the change. Due to this reason, once a high success rate is achieved it becomes increasingly hard to improve it further. On the other hand, improving a deterministic solver is a relatively easier task because all that is needed is to add new rules or change existing ones and test whether or not the success rates go up. Waiting for the lengthy training phase makes the improvement of learners a hard task.



    \chapter{Conclusions}

\section{Outcome}

Based on the findings of the \nameref{ch:evaluation} chapter, it can be concluded that it is wrong to assume that a highly efficient learner for Minesweeper, which models Minesweeper by only playing it, cannot exist. The answer to the central question of this research is that we can create efficient learners having performance close to that of the deterministic solvers. Another outcome of the evaluation is that despite having high success rates, the learners always fell short the success rates of the deterministic solvers, usually by a small margin.

Training of the learners was the most time consuming overhead. Thanks to concurrently training multiple learners, this issue was tackled to some extent. Improving the success rates of the neural network learners proved to be a very challenging task. The primary reason was that after changing the parameters of a learner, the whole training needed to be done again, to see whether the new model improved, and this consumed critical time from the project.

It is also observed that the high success rates of the deterministic solvers have been reached after years of research in the best approaches for solving Minesweeper. On the other hand, there is a lack of research in machine learning methods for solving Minesweeper, especially the neural networks. Future work in the usage of neural network learners for playing Minesweeper could further improve the win rates.

When comparing the solvers, it is clear that CSP solver should be chosen for solving Minesweeper because it has slightly higher success rates and it beats the neural learner by a large margin when speeds are compared. However, it remains debatable whether or not machine learning approaches should be used when solvers for problems like Minesweeper are developed. A machine learning engineer may prefer developing machine learning approaches for such problems because they find it more fascinating and interesting. On the other hand, a typical computer scientist would argue that the problems that can be modelled by algorithmic approaches should be solved using hard coded rules based solvers. After experiencing how time consuming it is to create and improve highly accurate learners for Minesweeper, it can be concluded that the argument of the computer scientist is stronger.


\section{Final remarks}
All the aims and objectives of the project were achieved. According to the aims and objectives, different configurations and architectures of neural networks were trained and tested to select the best model for solving Minesweeper.

On a personal note, I am satisfied with the outcome of the project. I have learnt many lessons about deep learning and about many other technical aspects after this project. I can also say that this was a challenging project that required proper planning and time management for a successful completion. 

Evaluating my decisions at the end of the project, there are still some areas where I could have done better. For instance, I now think that choosing Python over Java would have been a better decision because Python is better known for deep learning, I also observed that Deeplearning4j library does not have many tutorials online as it is less known in machine learning community. After evaluating the learners, I now think that I should have given a higher priority to improving the best learner on the expert mode because this mode had the least amount of learners trained despite being the hardest one.

    \printbibliography[title=References]\addcontentsline{toc}{chapter}{References}
    
    
    
\end{document}